\documentclass{elsarticle}
\usepackage[tight,footnotesize]{subfigure}
\usepackage{url}

\journal{}

\begin{document}

\begin{frontmatter}

\title{Technical Report CMPC14-02:  Predictive Modelling of Bone Age through Classification and Regression of Bone Shapes}
 \author[UEA]{Anthony Bagnall}
\author[UEA2]{Luke Davis}

\address[UEA]{ajb@uea.ac.uk, University of East Anglia, Norwich, UK}
\address[UEA2]{University of East Anglia, Norwich, UK}


\begin{abstract}
Bone age assessment is a task performed daily in hospitals worldwide. This involves a clinician estimating the age of a patient from a radiograph of the non-dominant hand.

Our approach to automated bone age assessment is to modularise the algorithm into the following three stages: segment and verify hand outline; segment and verify bones; use the bone outlines to construct models of age. In this paper we address the final question: given outlines of bones, can we learn how to predict the bone age of the patient? We examine two alternative approaches. Firstly, we attempt to train classifiers on individual bones to predict the bone stage categories commonly used in bone ageing. Secondly, we construct regression models to directly predict patient age.

 We demonstrate that models built on summary features of the bone outline perform better than those built using the one dimensional representation of the outline, and also do at least as well as other automated systems. We show that models constructed on just three bones are as accurate at predicting age as expert human assessors using the standard technique. We also demonstrate the utility of the model by quantifying the importance of ethnicity and sex on age development. Our conclusion is that the feature based system of separating the image processing from the age modelling is the best approach for automated bone ageing, since it offers flexibility and transparency and produces accurate estimates.
\end{abstract}


\end{frontmatter}

\section{Introduction}
Bone age assessment typically involves estimating the age of a patient from a radiograph by quantifying the development of the bones of the non-dominant hand. It is used to evaluate whether a child's bones are developing at an acceptable rate, and to monitor whether certain treatments are affecting a patient's skeletal development. Currently, this task is performed manually using an atlas based system such as Greulich and Pyle (GP) \cite{greulich1959radiographic} or a bone scoring method like Tanner and Whitehouse (TW) \cite{tanner1975assessment}. Atlas methods such as GP involve comparing the query image to a set of representative hand radiographs taken from subjects at a range of ages. Scoring systems assign each bone to one of several predefined stages, then combine these stage classifications to form an age estimate.

Manual procedures are time consuming and often inaccurate. Automated systems for bone age assessment have previously been proposed. These either attempt to recreate the TW or GP methods \cite{ eord1993knowledge, mahmoodi2000skeletal, niemeijer2003assessing}, or construct regression models for chronological age \cite{adeshina-evaluating, thodberg2009bonexpert}. Our approach is modular and feature based, and can be used to either recreate TW scores or predict age directly.

To predict TW bone stages we train a range of classifiers on three transformations of the outline. The first uses an ensemble technique described in~\cite{lines14elastic} that uses elastic distance measures directly on a one dimensional representation of the bone outline. The second technique finds discriminatory subsequences of the one dimensional series (called shapelets) through a transformation described in~\cite{hills13shapelet} and constructs classifiers in the shapelet feature space. Finally, we derive a set of summary shape features based on the TW descriptors. We conclude that the classifiers built on shape features are significantly better on at least one bone and provide greater explanatory power.

To predict age directly, we perform linear and non linear regressions from the shape feature space to age. We evaluate this process on a data set of images taken from~\cite{cao2000digital} in the age range 2--18. We show that, given the correct outline, we can accurately recreate TW stages and, using just three bones, can predict chronological age as accurately as clinical experts.

This stepwise, feature driven approach to automated bone ageing is transparent and explicable to clinicians. By separating out the feature extraction from the segmentation and regression we retain the potential for quickly and simply constructing new models for regional populations. This offers the possibility of producing age estimates tailored to local demographics based on data stored locally in film free hospitals.

The rest of this paper is structured as follows. In Section \ref{sec:BAA}, we review the current manual methods, and describe previous attempts at automated bone age assessment.

In Section \ref{sec:extraction} we describe how we format the segmented bones into outlines and shape features and in Section~\ref{classifiers} we provide an overview of the classification and regression techniques we use to predict bone age. In Sections~\ref{sec:TWStages} and \ref{regression} we present our results. Finally, we discuss our conclusions and describe the future direction of this work in Section~\ref{sec:conclusions}.

\section{Background}
\label{sec:BAA}
Bone age assessment is a task performed in hospitals worldwide on a daily basis.  The skeletal development of the hand is most commonly assessed using one of two methods: Greulich and Pyle (Section \ref{sec:gp}) \cite{greulich1959radiographic} or Tanner and Whitehouse (Section \ref{sec:tw}) \cite{tanner1975assessment}. The bone age estimate obtained by one of these methods is compared with the chronological age to determine if the skeletal development is abnormal. If there is a significant difference between the patient's bone age and chronological age then the paediatrician may, for example, diagnose the patient with a disorder of growth or maturation~\cite{hsieh2007bone}.

\subsection{The Greulich and Pyle Method}
\label{sec:gp}

The Greulich and Pyle (GP) method uses an atlas of representative hand radiographs taken from subjects at a range of ages. The latest (second)  edition of the GP atlas was released in 1959 \cite{greulich1959radiographic} which included new images for four new age points. The final atlas consists of 31 standard radiographs of males from newborn to the age of 19 years and 27 standard radiographs of females from newborn to the age of 18 years. Along with each standard, there is a piece of text describing the development. To use the atlas, the clinician checks the patient's radiograph against each of the example radiographs of the appropriate sex. The key features to check are the development of the epiphysis (the region at the end of particular bones) and the presence of certain carpal bones. The age estimate is the age of the subject who provided the representative image selected as the closest by the clinician.

This process is clearly somewhat subjective and a large variation between clinicians has been observed \cite{bull1999bone}. The representative images are from a very restricted sample taken over 60 years ago, and variation, changes in diet, healthcare and culture may mean this sample is no longer representative. Another criticism of GP is that the method implies an assumption that the ossification process happens in an linear fashion, which may not be true.

\subsection{The Tanner and Whitehouse Method}
\label{sec:tw}

In 1975, Tanner {\em et al.} \cite{tanner1975assessment} published a scoring system commonly referred to as TW2. Two separate methods of calculating bone age are described. The first method uses the radius, ulna and short bones (RUS) (the short bones cover the metacarpals and phalanges of fingers one, three and five). The second method uses just the carpal bones. The RUS method has been found to outperform the carpal bone technique and is easier to use.  Each bone has various stages associated with it and each stage has certain descriptors to use in the classification. Table \ref{tab:twstage} shows an image of each stage for the distal phalange of the middle finger with the associated criteria. Once all bones have been awarded a score, these scores are summed to find the Skeletal Maturity Score (SMS). The  distribution of ages for the SMS is described by a centile chart, from which a point estimate of bone age can be derived. The TW3 method was published in 2001 \cite{tanner-assessment}.  The basic maturity stages and scores remained the same but the centile charts have been updated to adapt to the modern population.

\begin{table}[!htb]
\caption{The various TW stages of Distal Phalange three \cite{tanner1975assessment}}
\scriptsize
\label{tab:twstage}
\begin{center}
\begin{tabular}{c|c|p{5cm}}
\hline
\textbf{Stage} &\textbf{Image} & \textbf{Description} \\
\hline
\hline
\textbf{B} & \raisebox{-.8\height}{\includegraphics[width=0.5cm]{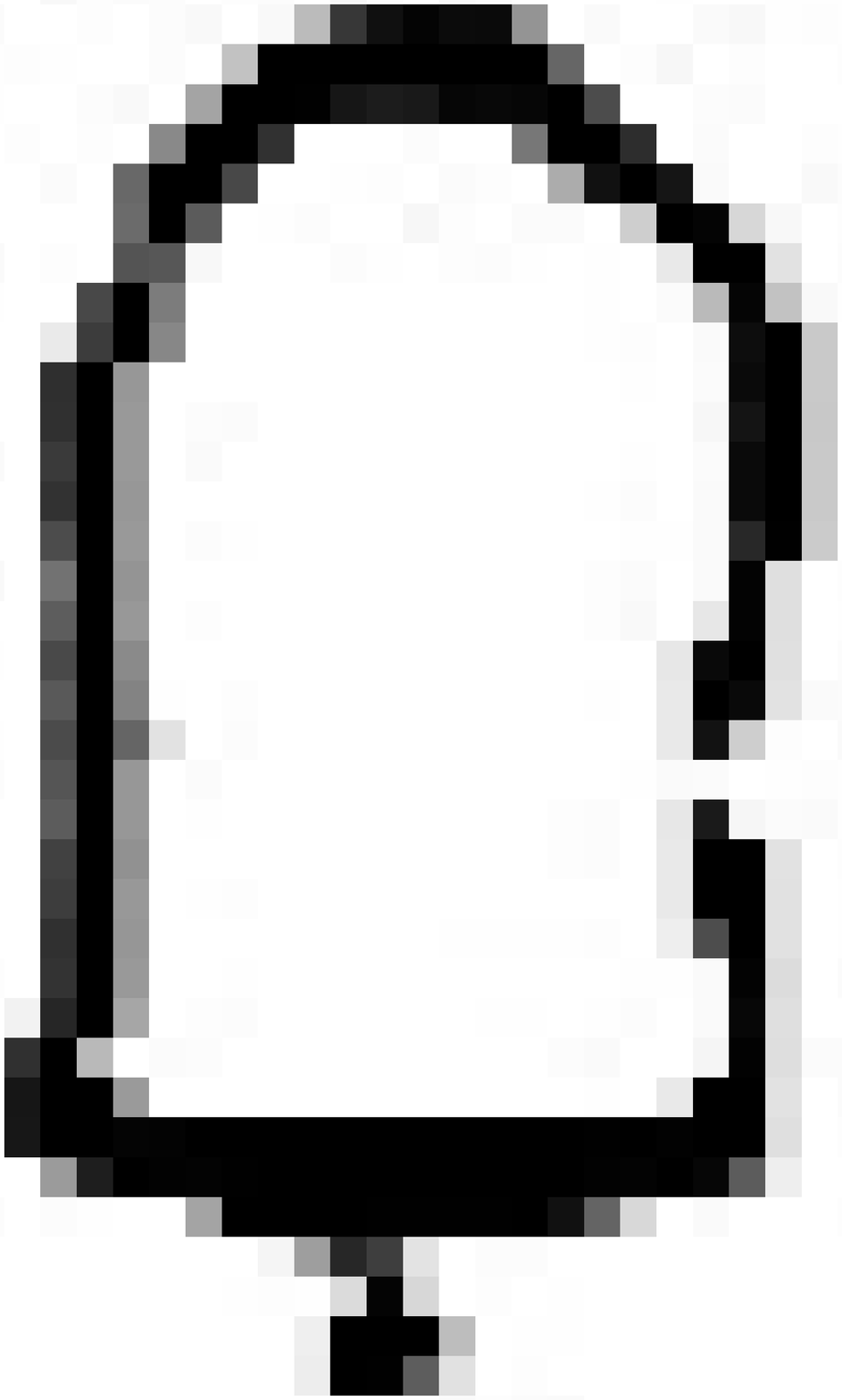}} & The centre is just visible as a single deposit of calcium, or more rarely as multiple deposits. The border is ill-defined.\linebreak\\
\textbf{C} & \raisebox{-.8\height}{\includegraphics[width=0.5cm]{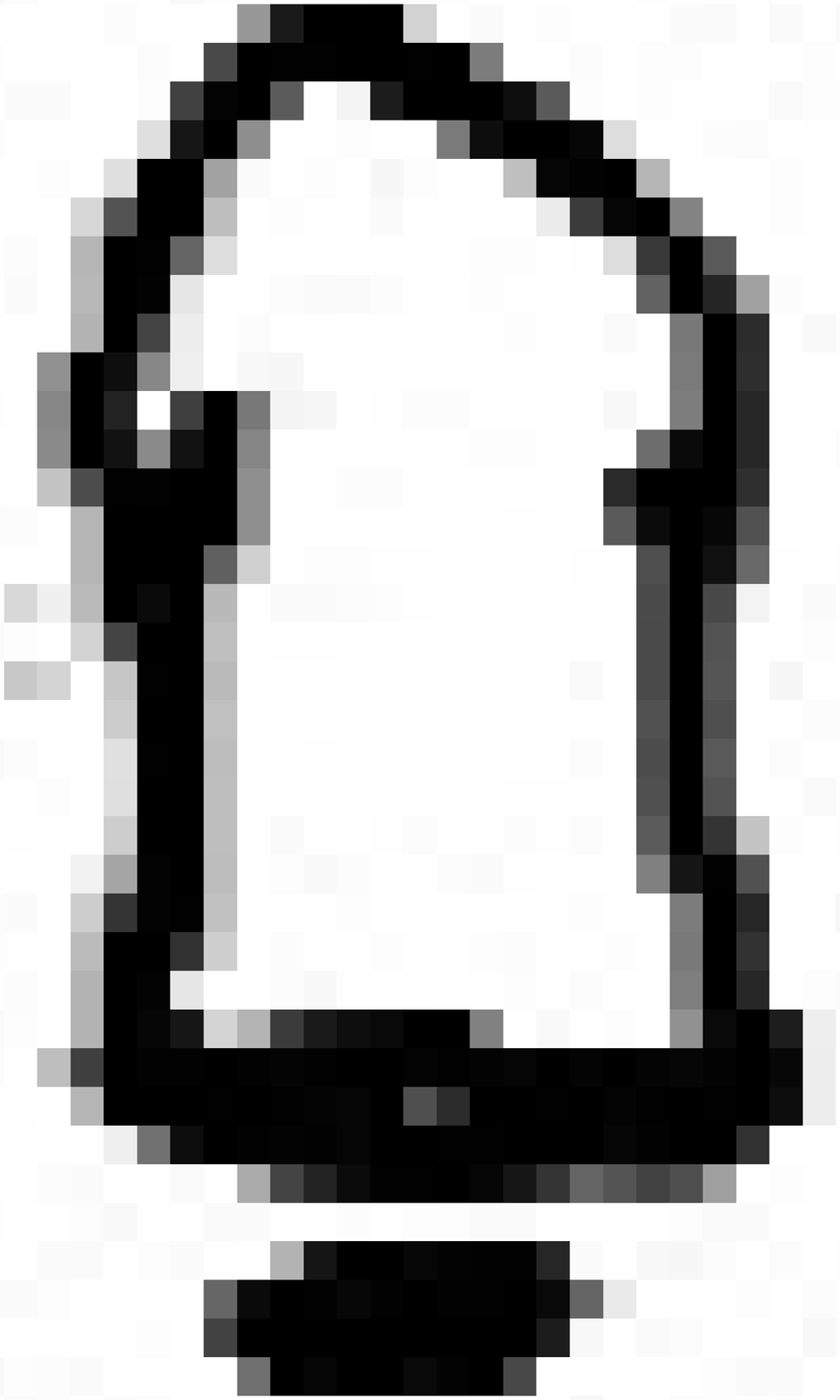}} & The centre is distinct in appearance and disc-shaped, with a smooth continuous border. \linebreak\\
\textbf{D} & \raisebox{-.8\height}{\includegraphics[width=0.5cm]{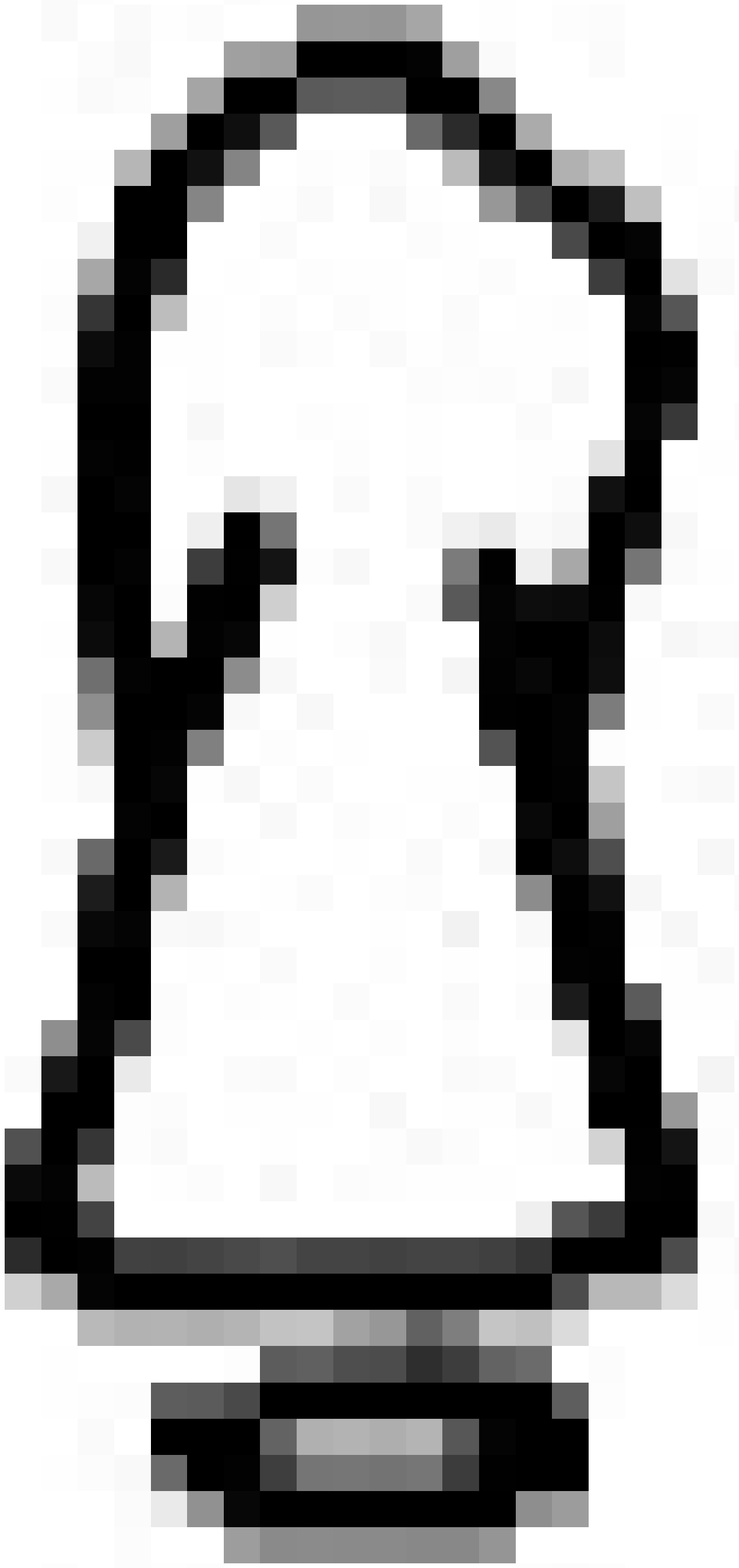}} & The maximum diameter is half or more the width of the metaphysis.\linebreak\\
\textbf{E} & \raisebox{-.8\height}{\includegraphics[width=0.5cm]{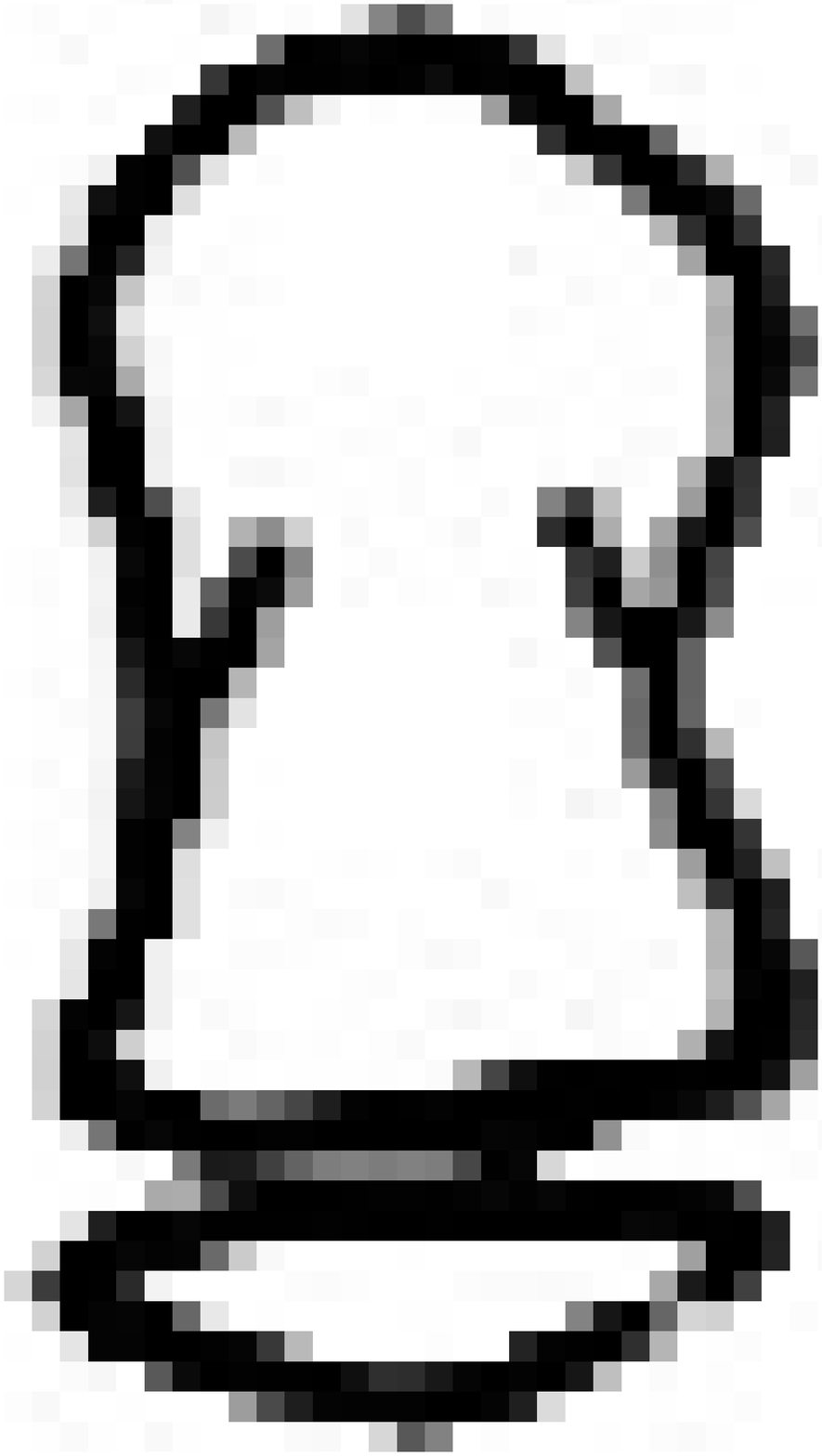}} & The epiphysis is as wide as the metaphysis. The central potion of the proximal border has grown toward the end of the middle phalanx, so that the proximal border no longer consists of a single convex surface; no differentiation into palmar and dorsal surfaces, however, can yet be seen.\linebreak\\
\textbf{F} &\raisebox{-.8\height}{ \includegraphics[width=0.5cm]{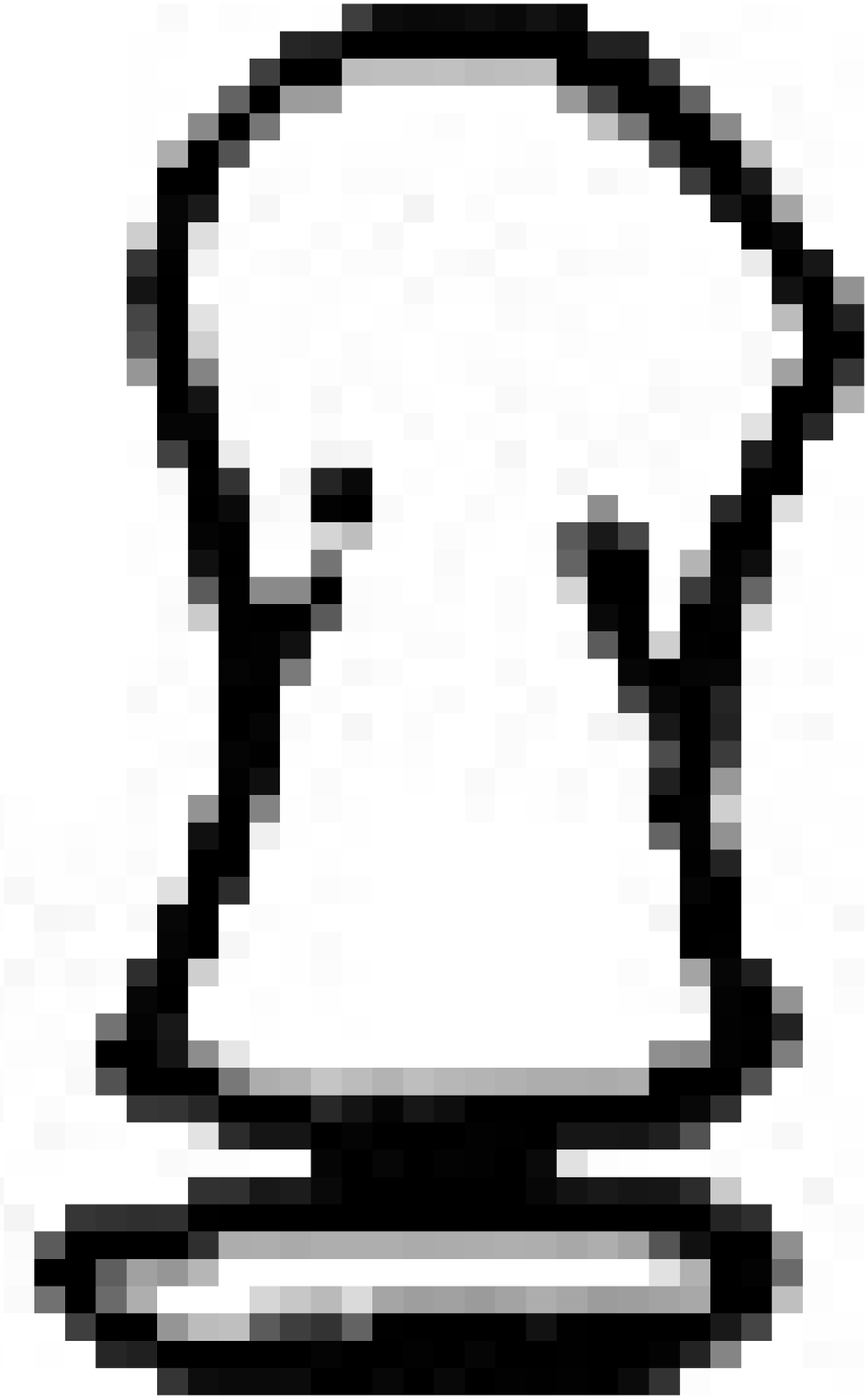}} & Palmar and dorsal proximal surfaces are distinct, and each has shaped to the trochlear articulation of the middle phalanx. The palmar surface appears as a projection proximal to the thickened white line representing the dorsal surface.\linebreak\\
\textbf{G} &\raisebox{-.8\height}{ \includegraphics[width=0.5cm]{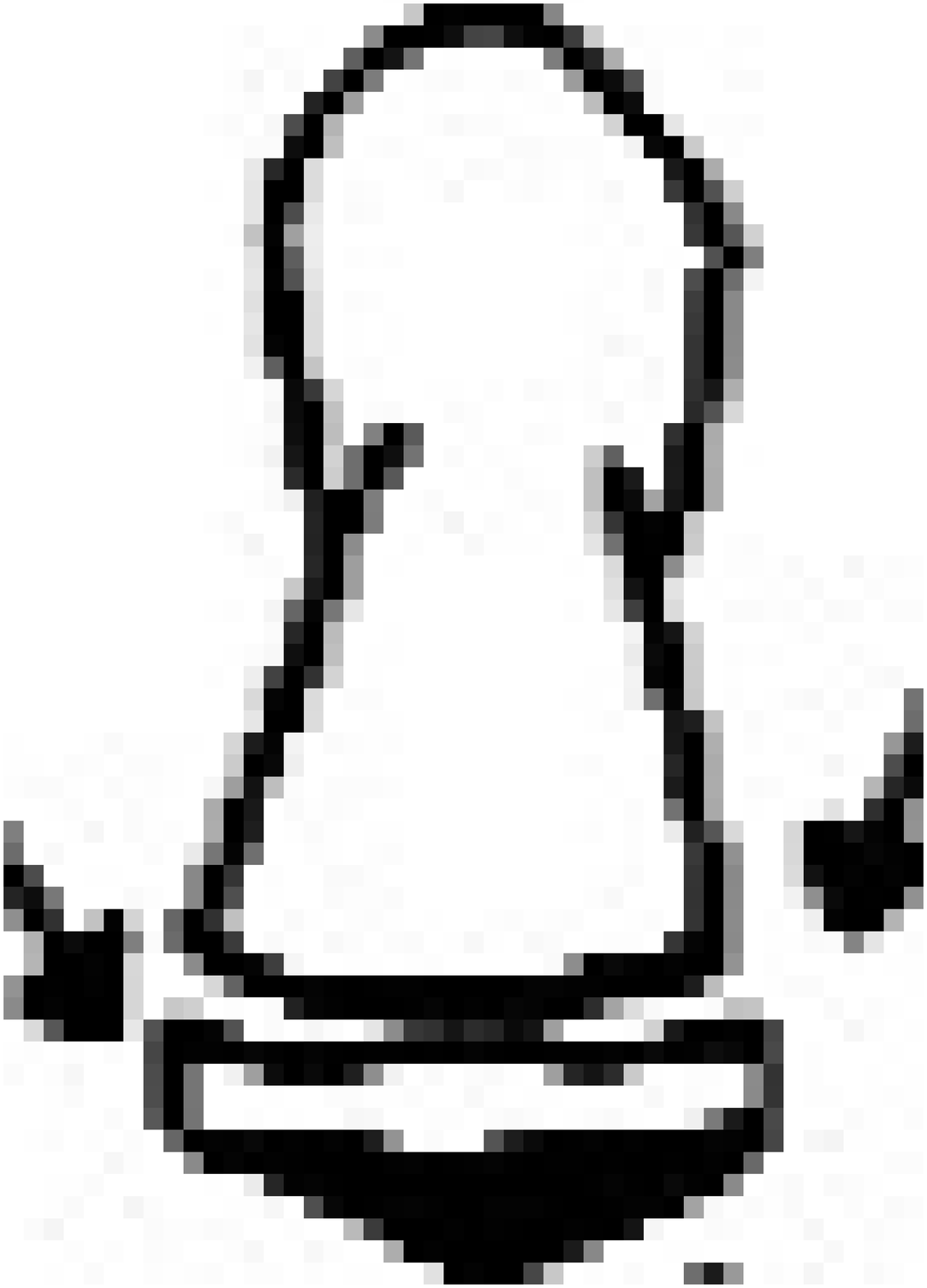}} & The epiphysis caps the metaphysis. \linebreak\\
\textbf{H} & \raisebox{-.8\height}{\includegraphics[width=0.5cm]{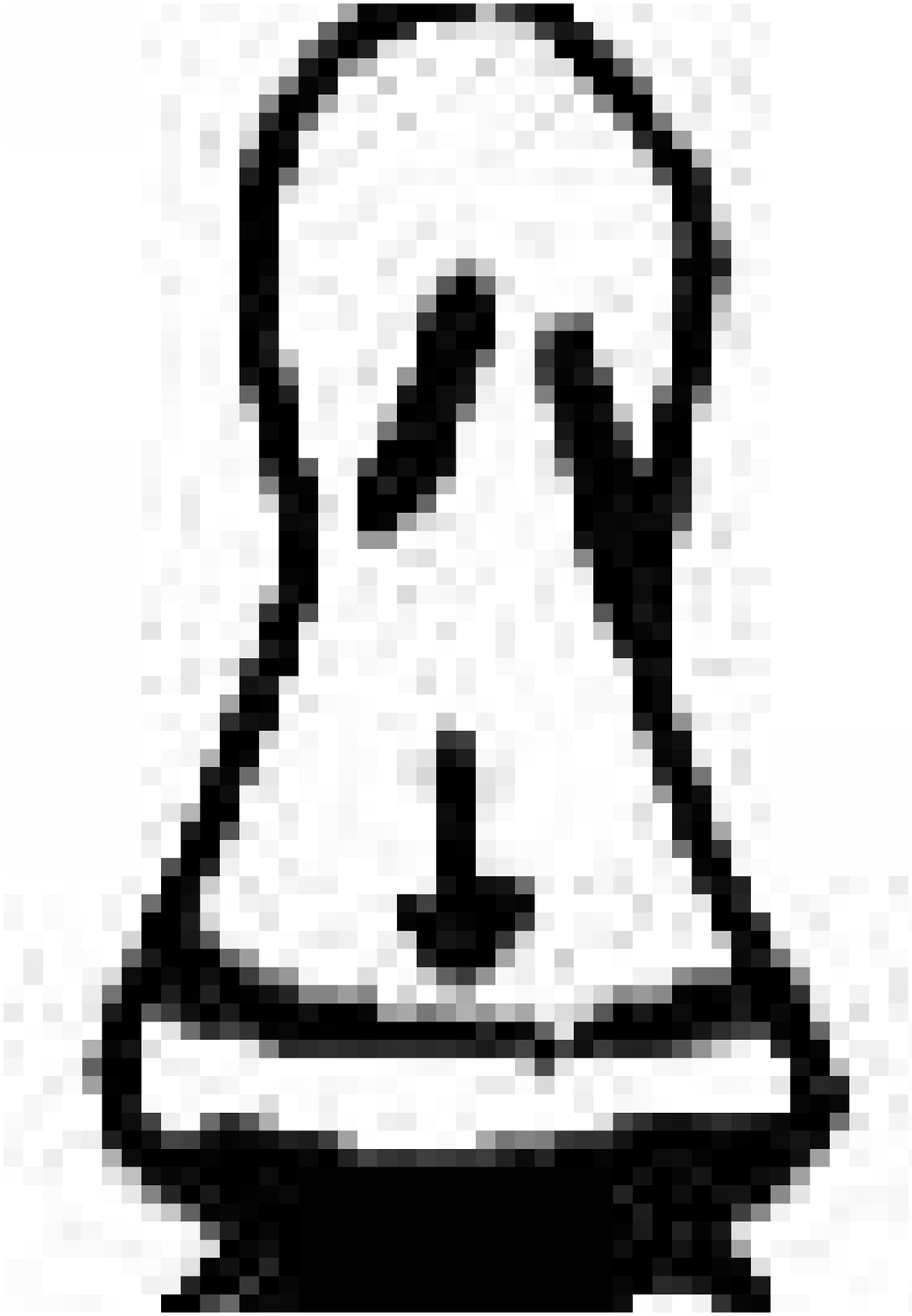}} & Fusion of epiphysis and metaphysis has now begun.\linebreak \\
\textbf{I} & \raisebox{-.8\height}{\includegraphics[width=0.5cm]{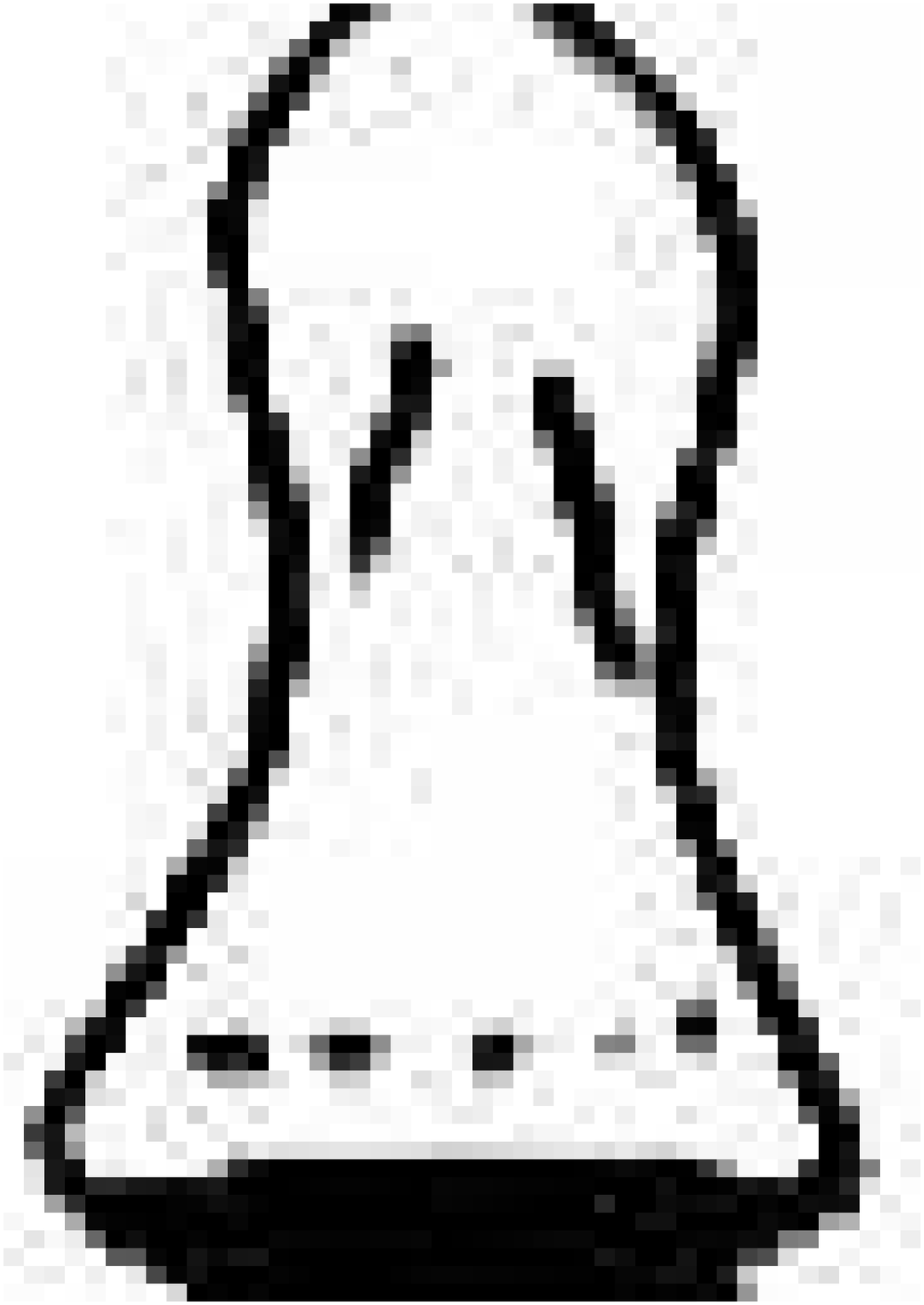}} & Fusion of epiphysis and metaphysis is completed.\linebreak\\
\hline
\end{tabular}
\vspace{-0.5cm}
\end{center}
\end{table}

The advantages of using TW in comparison to GP are that it is more objective and therefore results are more reproducible \cite{bull1999bone}. It does not have a strict order of ossification. A centile chart can be calculated for any population which makes the method adaptable to local conditions. The TW method does not have discrete intervals and therefore removes some of the restrictions of the GP method. However, performing TW scoring is time consuming, and given the pressure on clinicians time, the GP method is used more often.

\subsection{Related Automated Bone Ageing Systems}
\label{sec:auto_systems}
Thodberg {\em et al.} \cite{thodberg2009bonexpert} use Active Appearance Models (AAMs)~\cite{cootes2001active} for their automated bone age assessment system (BoneXpert). The proposed algorithm consists of three layers: A, B and C. Layer A involves fitting separate AAMs to each of the 15 RUS bones separated into three separate age epochs (3 to 8.2 years, 8.2 to 13 years, and 13 to 18 years for boys). The BoneXpert bone age is constructed in layer B directly from the models constructed in layer A. A linear regression of 10 shape, 10 intensity and 10 texture features onto chronological age for each short bone and epoch combination is performed. The bone age estimates are averaged over the bones within each epoch, and this average bone age estimate is used as the independent variable for a further regression for each bone/epoch combination. This second regression is the final model for each bone/epoch combination. Layer C involves fitting and using these models on previously unseen data. The authors present results comparing their age estimates to the images in the GP atlas and describe and evaluate a mechanism for recreating TW scores. When validated against the images in the GP atlas, BoneXpert estimates $B_1$ are on average 0.7 years larger than the actual ages of the subject. The authors then perform a post hoc adjustment by subtracting 0.7 from each estimate. They justify this by citing differences in populations between the GP atlas subjects and those used to construct the model. After this adjustment, the BoneXpert predictions have standard deviation of 0.42 years to the true age of the GP atlas ages. To recreate TW, the bone age estimates are mapped onto the TW scores using a training set of images with TW ratings assigned by a human operator. On a cross validation of 84 radiographs they report 68\% agreement between BoneXpert and the human scorer, with 94\% of the estimates within one stage of each other. For the phalanges, BoneXpert is in agreement with the human on approximately 70\%-80\% of bones.

Adeshina {\em et al.} \cite{adeshina-evaluating} describe another system built on AAMs. 170 images from patients between 5 and 20 years of age (87 male and 83 female) were manually annotated with 330 landmark points which were then used to fit a more detailed shape using a non-rigid registration algorithm. An AAM model is trained on the whole data set and a linear regression model from the AAM features to chronological age fitted. Different regression models are used for male and female patients. The paper compares the difference between single AAMs for each bone and combined sets of bones, e.g. the carpals. The combined models slightly outperform the individual bone models. Using leave one out cross validation, the average mean absolute error for the single bone models was $1.47\pm0.08$ years against chronological age for females and $1.26\pm0.07$ for males. The reported performance of the models using the 13 RUS bones had mean absolute errors of $0.80\pm0.09$  and $0.93\pm0.08$ for females and males respectively. The algorithm is essentially a simpler version of BoneXpert (with the addition of a registration phase) that relies on an AAM for the outline and feature extraction. Whilst the age estimation is evaluated on unseen data through cross validation, there is no discussion as to the accuracy of the outline detection algorithm on unseen data.

Niemeijer {\em et al.}  \cite{niemeijer2003assessing} outline a method to automate skeletal age assessment that uses Active Shape Models (ASMs)~\cite{cootes2001active} to segment the distal phalanx of the third finger. They construct a separate model on training data for each TW stage (E--I). For new data they use these models to extract the phalanx, then measure the similarity of the fitted bone to the training bones in the model space, with a nearest neighbour, maximum correlation and linear discriminant approach. They evaluate their system on 71 images by comparing the predicted TW stage against the TW ratings of two clinicians.  The results show the second clinician gave the same stage as the first clinician on 80.3\% of the rated bones and within one stage 100\% of the time. The results presented from the proposed system assigned 73.2\% of bones with the correct TW stage and 97.2\% within one TW stage.

Efford's \cite{eord1993knowledge} method for automatically assessing skeletal maturity uses ASMs to segment the bone. Firstly, a hand silhouette is generated by thresholding the image to provide a binary mask, which is then filtered using various morphological operators. A vertical line that intersects the phalanges of the middle finger and a horizontal line that intersects the metacarpals are calculated. This helps locate the other bones in the hand and wrist. The shape of the silhouette is analysed to ensure that it is as expected for a normal hand. This is done using chain codes with 11 landmarks: five fingertips, four between fingers and two for the wrist. The bones are segmented using ASMs in the following order: radius, ulna, metacarpals, carpals, phalanges. To assess the maturity, each criteria from the TW2 stages is changed into code, features are extracted from the test image and compared to see if the criteria for a TW2 stage are met e.g. metaphysis and epiphysis are the same width. No empirical results of the proposed system were presented.

In ~\cite{pietka2001computer,pietka2003integration}, Pietka {\em et al.} describe a method that uses \textit{c}-means clustering and Gibbs Random Fields to segment bones from a radiograph. Six regions of interest (ROI) are located: the joints between distal phalanges - middle phalanges and middle phalanges - proximal phalanges of fingers two, three and four. These ROIs are segmented and features are extracted from them. Firstly, background subtraction is performed using histogram analysis, then the axes of fingers two, three and four are located. These axes are used to locate the six ROIs. The detection of ROIs is found to be between 84\%-95\% accurate.  The bone is extracted from the soft tissue in the ROIs and a variety of features describing the bones are derived using wavelet decomposition. The  discriminatory power of these features is assessed through plots against patient age.

Bone ageing consists of three stages: locating the relevant bones; deriving discriminatory features from the bones; and regressing these features onto age (or constructing a classifier to recreate TW stage). The majority of research in this field has used ASMs or AAMs to combine the first two stages of locating the bones and deriving features. We have experimented with ASMs/AAMs~\cite{davis2012segmentation}, but have ultimately rejected the approach for a methodology that is closer to that of Pietka {\em et al.} ~\cite{pietka2001computer,pietka2003integration}. Whilst AAMs obviously can perform hand / bone segmentation well, there were several reasons for not pursuing this approach. Firstly, AAMs require manually labelling ``landmark" points on a training set of images. The bones of the hand are fairly simple shapes that do not have many natural landmarks and hence the placement of landmarks can be highly variable between subjects. Secondly, using the model to segment a new image requires a starting template close to the correct position. We found that even with the hand outline the AAM would fit a hand shape that was in fact an outline of the carpal bones. This tendency to fit a valid shape in the wrong location makes automated validation of the process difficult. Thirdly, the requirement of training data means that the model is only representative of the population from which the training data is sampled. This makes it hard to develop models tailored to specific demographics without labelling a whole new sample of images. Fourthly, the standard use of AAMs is to capture variation with a homogeneous population in order to use this to detect whether new images are outliers or members of a different population. With hand images, there is a wide variation between the members of the population, and the variation is continuous. BoneXpert overcomes this by splitting the population into three age groups, but this requires three times as much training data and introduces complexities into the predictive stage. Finally, the features the AAM derives do not necessarily have any direct clinical interpretation, and hence make it harder to use the model to explain the relationship between physical characteristics and age estimates.

\section{Bone Outline Transformation}
\label{sec:extraction}

\begin{figure}[htb]
 		\begin{center}
		\includegraphics[height = 6cm]{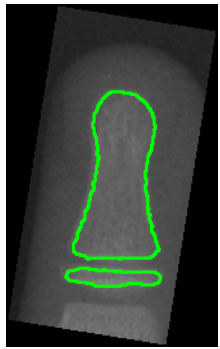}
		\caption{An example segmentation of the proximal phalanges TW stage F.}	
		\end{center}
\label{outline}
\end{figure}

The Automated Skeletal Maturity Assessment (ASMA) algorithm described in~\cite{davis13phd} consists of the following distinct stages: segment and verify hand outline; segment and verify bones; and use the bone outlines to construct models of age. The first two stages are described in~\cite{davis2012segmentation,davis2012assessment,davis13phd}. For the third stage, we assume the location and outline of the bone have been correctly identified. We use radiographs taken by the Childrens Hospital Los Angeles~\cite{cao2000digital}. The age range of subjects is 2--18. Restriction to this age group is common with automated bone ageing~\cite{adeshina-evaluating, thodberg2009bonexpert} since it has been shown that bone age assessment is unreliable on radiographs from patients under the age of 2 years~\cite{bull1999bone}. The images have two clinicians GP ratings associated with them~\cite{gertych2007bone}, which we use in bench mark comparison in Section~\ref{prediction}. The dataset also includes labels for four ethnicities: Asian, African-American, Caucasian and Hispanic, which we use to demonstrate the utility of the approach in Section~\ref{exploration}. The question we wish to address in this paper is how best to use a correctly located bone in classification and regression models. Hence, we assume that a correct outline of the three bones of the middle finger have been extracted. Figure~\ref{outline} shows an example of a correct outline segmentation. Details of how this is obtained are given in~\cite{davis13phd}.

A prerequisite to constructing predictive models is deciding on how best to represent the bone to work with a classifier. There are three possible approaches. We could extract shape and/or intensity features (such as 2-D Fourier transforms) directly from the bone outline or image intensity. Alternatively, we may extract the image outline and build models based on the outline characteristics. Thirdly, we can derive summary shape features based on the TW descriptors. Our preliminary experiments with the first approach were not promising, due to the wide variation of intensity distributions between images. We concentrate on using the outline directly (Section~\ref{1d}) and deriving descriptors of the outline (Section~\ref{feature}).

\subsection{1-D Outline Representation}

\begin{figure}[htb]
 		\begin{center}
\begin{tabular}{cc}
		\includegraphics[width = 3cm, height = 4.5cm]{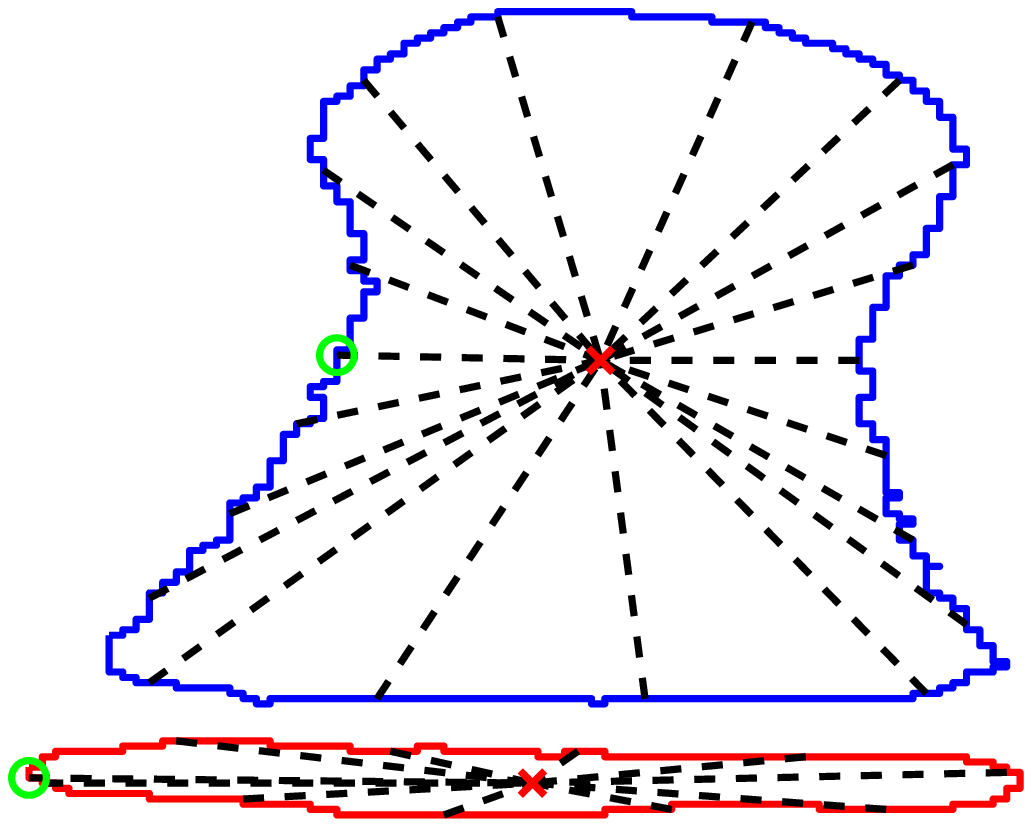}
&
		\includegraphics[height = 4.5cm]{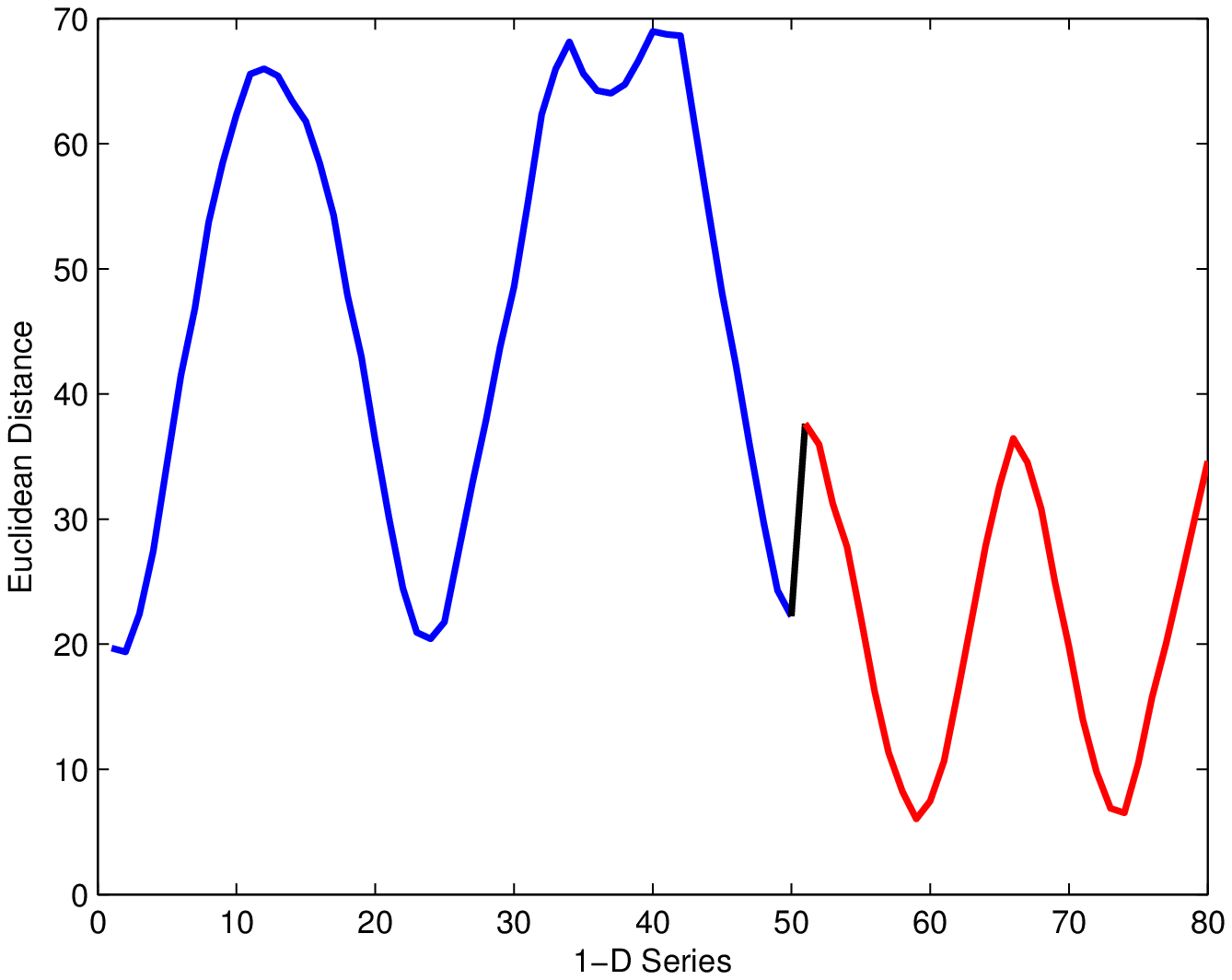}
\end{tabular}
		\caption{An example mapping of a 2-D outline to a 1-D series of distances to centre.}	
		\end{center}
\label{outline2}
\end{figure}

The one-dimensional series is obtained by calculating the Euclidean distance of each pixel along the outline of the bone. We align the 1-D series by using the midpoint of the phalanx as the starting point of the series and move around the hard tissue in a clockwise direction. If an epiphysis is present the series is concatenated to the phalanx series. Clearly, the length of outlines will vary. To simplify the classification the outlines were resampled to ensure each was the same length as the shortest series (80 attributes, 50 phalanx, and 30 epiphysis). If no epiphysis is present the phalanx series is followed by 30 zeros. An example of this process is shown in Figure~\ref{outline2}.

\label{1d}

\begin{figure}[htbp]
	\centering
\begin{tabular}{cc}

       \includegraphics[width=5.5cm]{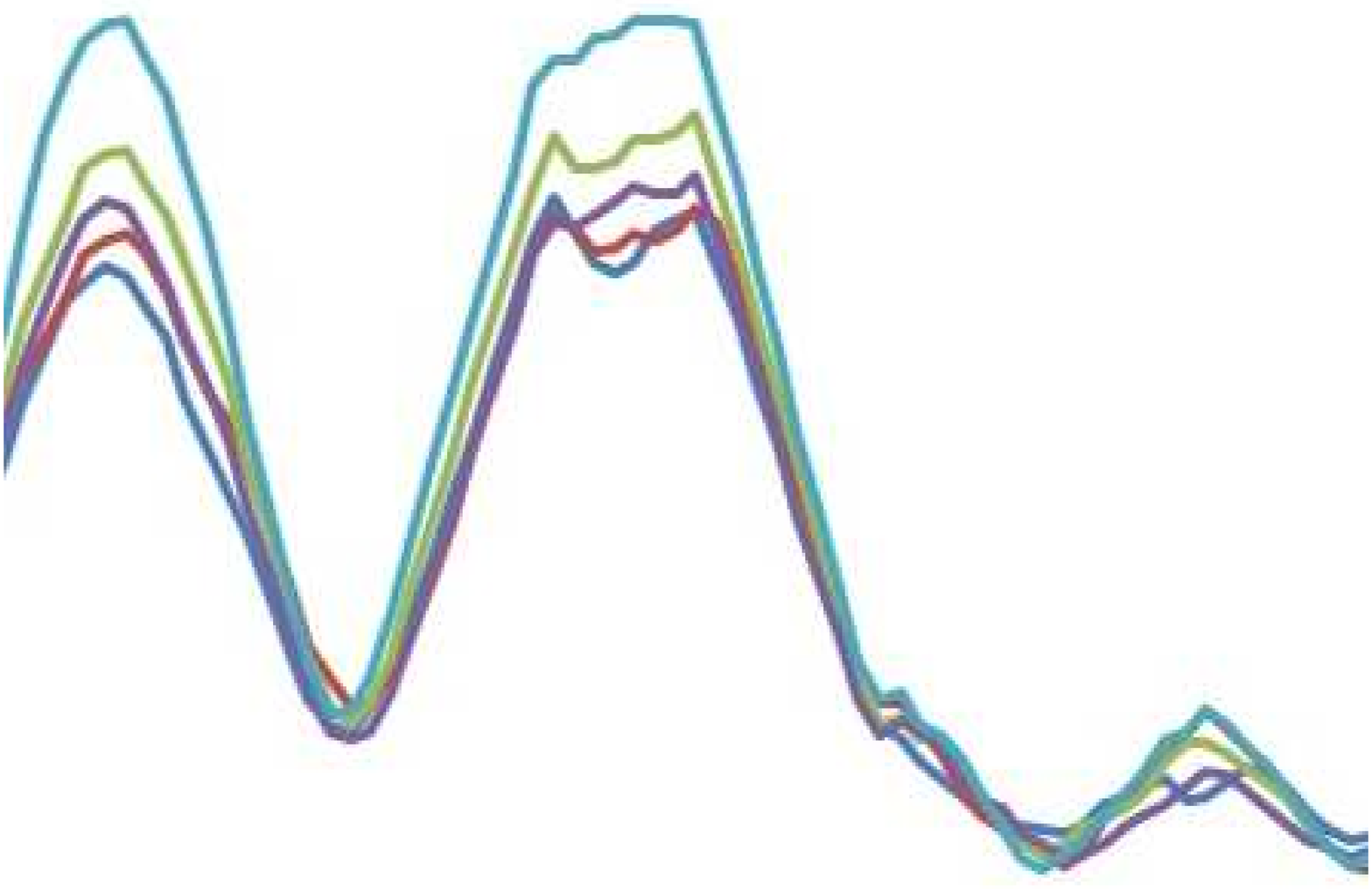}
&
       \includegraphics[width=5.5cm]{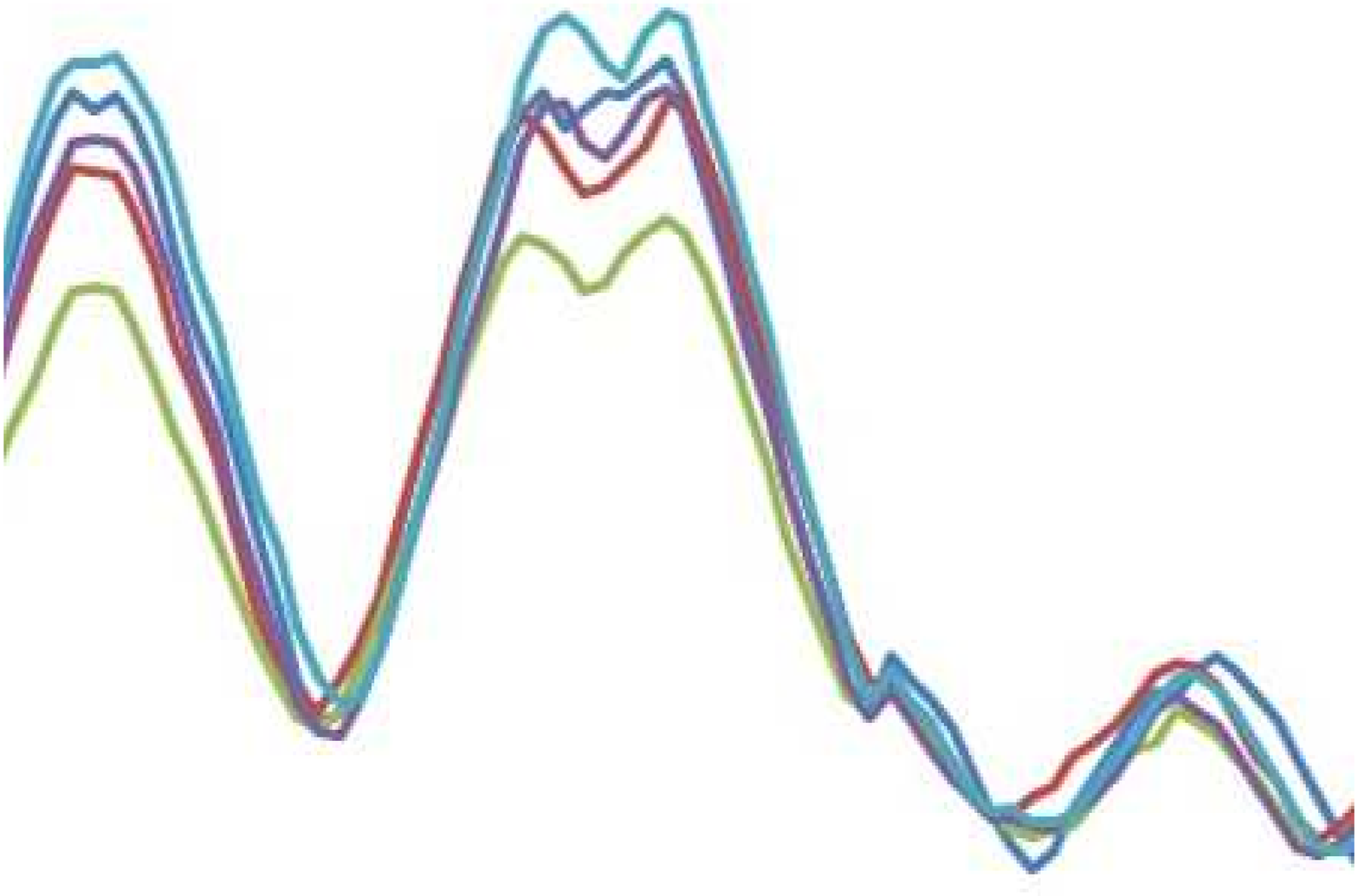}
\\
Stage D  & Stage E\\
       \includegraphics[width=5.5cm]{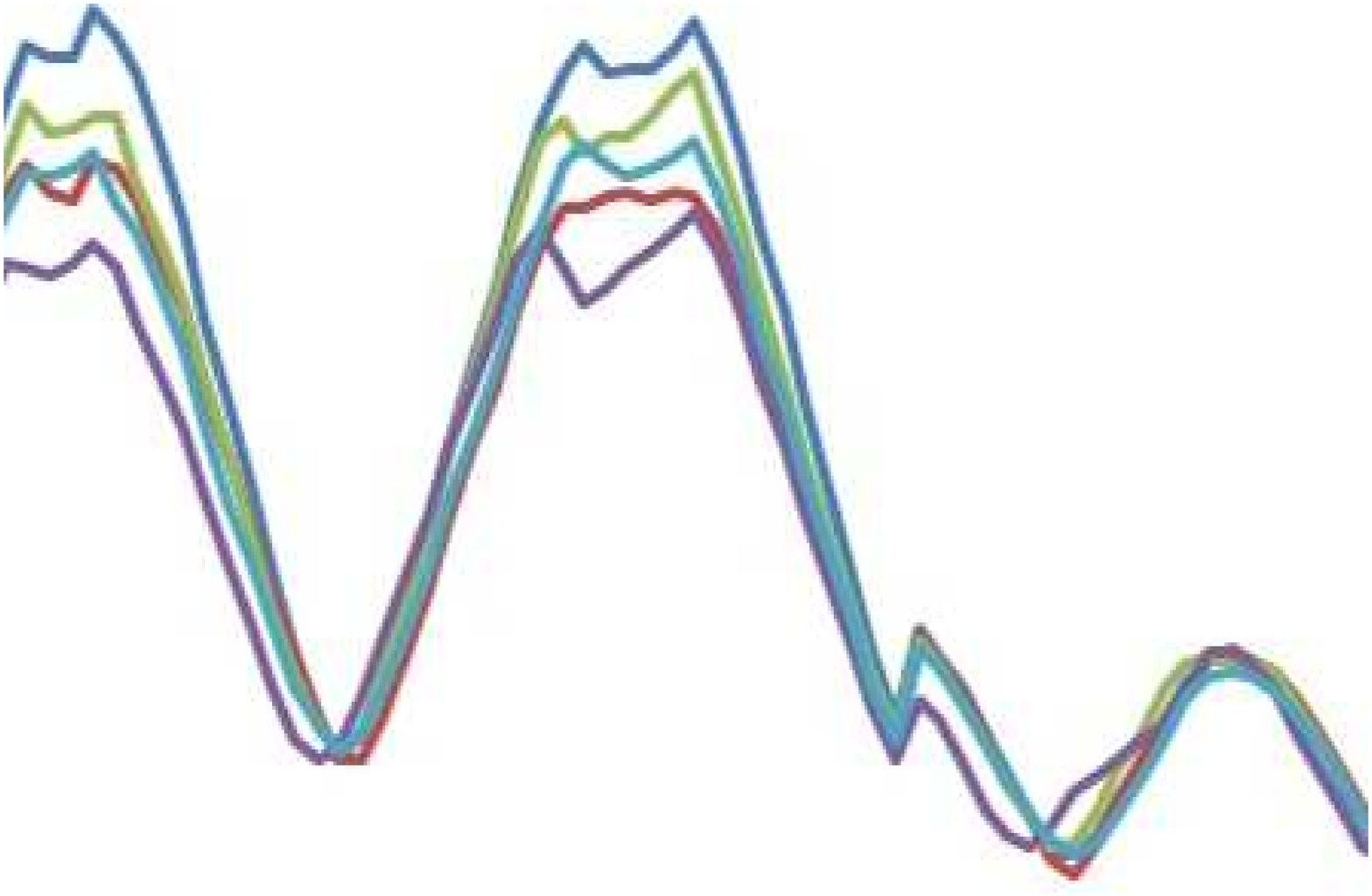}
&
       \includegraphics[width=5.5cm]{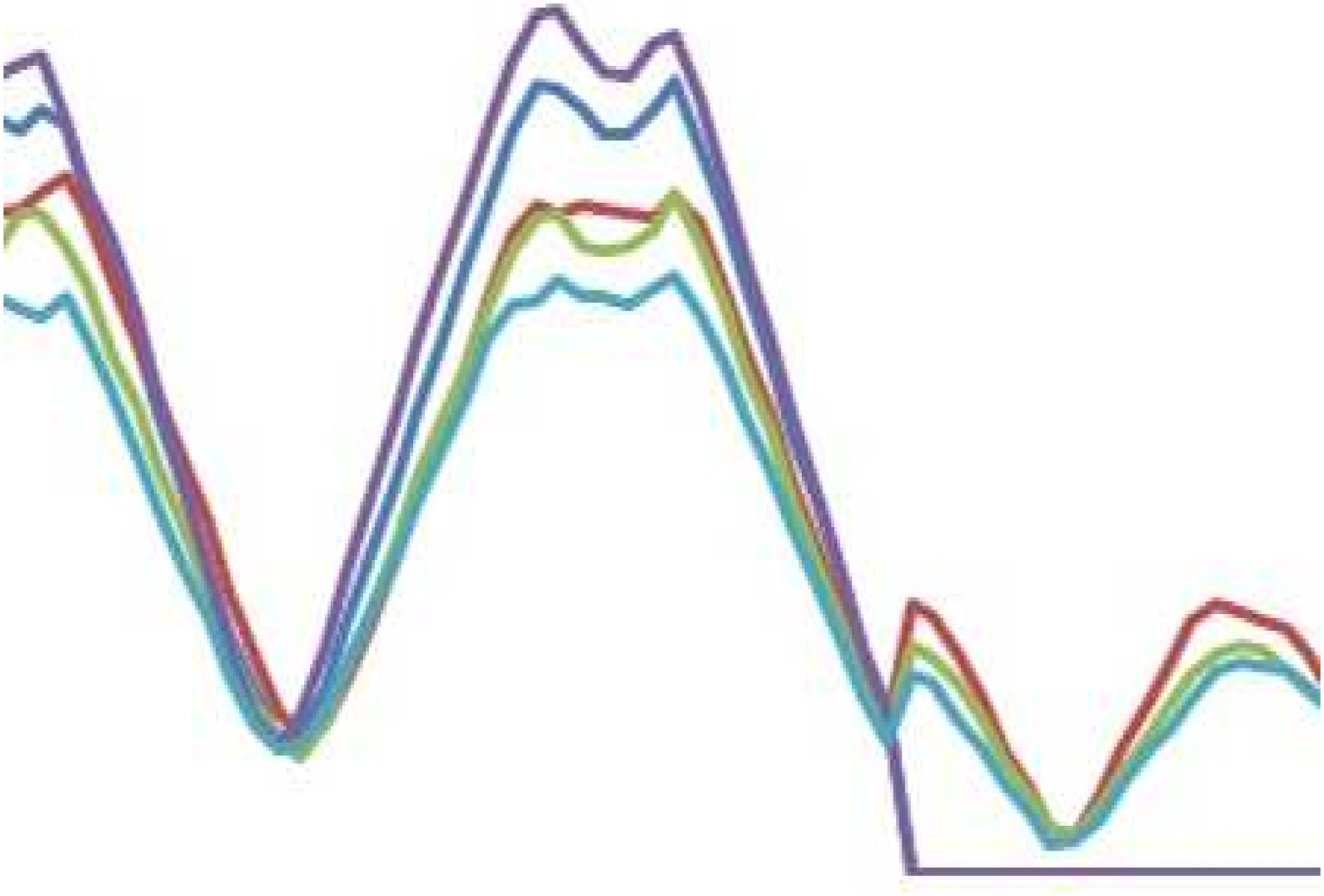}
\\
Stage G  & Stage H\\
\end{tabular}

	\label{example1D}
\end{figure}

The premis of this approach is that the shape variation between TW stages will be detectable in the 1-D series by utilising the myriad of algorithms developed in time series data mining. Figure~\ref{example1D} shows examples of series from four different TW stages. Some differences are obvious and should be easily detected. Some Stage H bones and all Stage I bones have no epiphysis, and so have a flat line tail. Other differences are more subtle. For example, there is a double dip in the first peak for stages G and H, but not for stage D and E.

\subsection{Shape Features Representation}
\label{feature}

The features we extract are described in Table~\ref{tab:features}. Our choice of features is based on the text for classifying TW stages (see Table~\ref{tab:twstage}).

\begin{table}[!ht]
\caption{Features Extracted From Segmented Bones}
\scriptsize
\label{tab:features}
\begin{center}
\begin{tabular}{ c | l }
\hline
\textbf{Feature Number} &\textbf{Feature Name} \\
\hline
\hline
\textbf{1}& Epiphysis Present \\
\textbf{2} & Phalanx Ellipse Height \\
\textbf{3} & Phalanx Ellipse Width \\
\textbf{4} & Phalanx Height \\
\textbf{5} & Phalanx Width \\
\textbf{6} & Phalanx First Quartile Width \\
\textbf{7} & Phalanx Third Quartile Width \\
\textbf{8} & Metaphysis (Phalanx  Ninety Percentile) Width \\
\textbf{9} & Phalanx Eccentricity \\
\textbf{10} & Phalanx Width to Height Ratio \\
\textbf{11} & Phalanx Roundness \\
\textbf{12} & Phalanx Area to Perimeter Ratio\\
\textbf{13} & Phalanx First Quartile to Width Ratio\\
\textbf{14} & Phalanx Third Quartile to Width Ratio\\
\textbf{15} & Phalanx Metaphysis to Width Ratio\\
\textbf{16} & Epiphysis Ellipse Height \\
\textbf{17} & Epiphysis Ellipse Width \\
\textbf{18} & Epiphysis Height \\
\textbf{19} & Epiphysis Width \\
\textbf{20} & Epiphysis Eccentricity \\
\textbf{21} & Epiphysis Distance to Phalanx \\
\textbf{22} & Epiphysis Width to Height Ratio \\
\textbf{23} & Epiphysis Roundness \\
\textbf{24} & Epiphysis Area to Perimeter Ratio \\
\textbf{25} & Epiphysis Width to Metaphysis Ratio \\
\hline
\end{tabular}
\end{center}
\end{table}

The most important feature is whether the epiphysis is present or not (feature 1). This can be calculated from the number of regions extracted during the segmentation. The other features are summary measures of the phalanx and the epiphysis (if present). The calculation of these features is described in detail in \cite{davis2012assessment}.

We extract basic size descriptors such as height and width as these should be indicative of age. However, the size of the image is not highly correlated to the size of the hand since the focus of the x-ray machine is adjusted to standardise the hand size. The obvious way to find the height and width of the phalanx and epiphysis is to find the length of the vertical line down the centre of the bone for height and the length of the horizontal line across the middle of the vertical for width. However, this assumes the bones are vertically aligned, which is often not the case, since fingers in radiographs are often not straight. In order to calculate an estimate for height and width, we fit an ellipse to both the phalanx and epiphysis (if present) using the Hough transform~\cite{ballard1981generalizing}.

We use the axes of the ellipse to calculate an estimate for the height and width of the phalanx and epiphysis (features 2, 3, 16 and 17). We also use the vertical axis of the ellipse to calculate the width of the phalanx at certain points along its length (features 5, 6, 7, 13 and 14). The shape of the phalanx and epiphysis both change during development, and we use two roundness measures to capture this (features 9, 11, 20 and 23).  The distance between the phalanx and epiphysis also changes over time, so we quantify this characteristic with the Euclidean distance between the midpoints of the two (feature 21).

\section{Classification and Regression Techniques}
\label{classifiers}

\subsection{Elastic Distance Measures}

The 1D outline series is commonly used directly in the times series data mining literature~\cite{UCRWeb} in combination with nearest neighbour classifiers and a specialised distance function. Potential localised misalignments are compensated for through some elastic adjustment in the distance function, such as variants of Dynamic Time Warping (DTW) and edit distance-based measures. It was shown in~\cite{lines14elastic} that significantly better classification accuracy could be achieved through combining these distance measures in an ensemble. The ensemble combines DTW, weighted DTW~\cite{jeong11weighted}, Longest Common Subsequence, Edit Distance with Penalty, Time Warp Edit Distance~\cite{marteau09stiffness} and Move-Split-Merge~\cite{stefan12msm}. The code is available from~\cite{SDM14}.

\subsection{Shapelet Transform}

One alternative to time domain classifiers is time series shapelets~\cite{ye09shapelets,hills13shapelet}. Shapelets are time series subsequences that are discriminatory of class membership. They allow for the detection of phase-independent localised similarity between series within the same class. The original shapelets algorithm by Ye and Keogh~\cite{ye09shapelets} uses a shapelet as the splitting criterion for a decision tree. Hills \textit{et al.}~\cite{hills13shapelet} propose a shapelet transformation that separates the shapelet discovery from the classifier by finding the top $k$ shapelets on a single run (in contrast to the decision tree, which searches for the best shapelet at each node). The shapelets are used to transform the data, where wach attribute in the new dataset represents the distance of a series to one of the shapelets. They demonstrate that the ability to use more sophisticated classifiers in conjunction with shapelets significantly reduces classification error. We use the code available from ~\cite{ShapeletWeb}.

\subsection{Classification Algorithms}

The classifiers used are the WEKA~\cite{hall2009weka} implementations of $k$ Nearest Neighbour (where $k$ is set through cross validation), Naive Bayes, C4.5 decision tree~\cite{quinlan1993c4}, Support Vector Machines~\cite{cortes1995support} with linear and quadratic basis function kernels, Random Forest~\cite{breiman2001random} (with 100 trees) and Rotation Forest~\cite{rodriguez2006rotation} (with 30 trees).

\section{Classification of Tanner-Whitehouse Stages}
\label{sec:TWStages}

Because we are performing extensive model selection, we split the data into train and test sets and performed all model selection on the training data. The number of instances for each bone is shown in Table~\ref{tab:tw_nums}.

\begin{table}[!ht]
\caption{Number of Instances for Tanner-Whitehouse Classification}
\scriptsize
\label{tab:tw_nums}
\begin{center}

\begin{tabular}{ r | c | c | c }
\hline
 & \textbf{No. of Instances} & \textbf{No. of Training} & \textbf{No. of Testing}\\
\hline
\hline
\textbf{Distal} & $539$ & $400$ & $139$ \\
\textbf{Middle} & $554$  & $400$ &$154$\\
\textbf{Proximal} & $605$ & $400$ & $205$\\
\hline
\end{tabular}
\end{center}
\end{table}

We conducted a ten fold cross validation classification experiment on the training data for each of the bones. We follow previous methodology~\cite{niemeijer2003assessing, thodberg2009bonexpert} by evaluating our models based on both the accuracy and the proportion correct within one TW stage.

Tables~\ref{tab:TWS_DP3},~\ref{tab:TWS_MP3} and~\ref{tab:TWS_PP3} show the results for a ten fold cross validation on the training data set. The shape feature and shapelet classifiers are constructed with WEKA default parameters. The elastic ensemble member parameters are set through cross validation.

These results demonstrate that, overall, the classifiers built on the shape features are more accurate than those constructed directly from the outlines or from shapelets. Secondly, the middle phalange is the hardest to accurately classify for all techniques, but especially for the elastic ensemble and the shapelet transform classifiers. This demonstrates that the outline is less discriminatory for this phalange, and TW class is determined more by the interaction of phalanx and epiphysis.

\begin{table}[!ht]
\caption{Cross validation classification accuracy (\%) for distal phalange III Tanner-Whitehouse stage on the training set.}
\label{tab:TWS_DP3}
\begin{center}
\scriptsize
\begin{tabular}{| r | c c | c c | c c |}
\hline
           & \multicolumn{2}{c|}{Shape Features} &  \multicolumn{2}{c|}{Shapelet Transform} & \multicolumn{2}{c|}{Elastic Ensemble}

            \\ \hline
Classifier          & \textbf{Correct} & \textbf{Within One} & \textbf{Correct } & \textbf{Within One} & \textbf{Correct } & \textbf{Within One}\\
\hline
\hline
 \textbf{NN}          & 81.00 & 97.00         &  74.50 &   92.75  &  81.75 & 96.50      \\
 \textbf{Naive Bayes}   & 80.00 & \textbf{98.75}& 72.50  & 93.75 & & \\
 \textbf{C4.5}          & 77.00 & 97.00         & 71.00 &  91.50 & & \\
 \textbf{SVML}          & 82.25 & 97.25         & 77.50 &  95.00 & & \\
 \textbf{SVMQ}          & \textbf{84.75} & 98.00 & 77.00 & 94.50 & & \\
 \textbf{Rand. Forest}  & 83.00 & 98.00         & 79.75 & 94.00& & \\
 \textbf{Rot. Forest}   & 81.25 & 97.75         & 77.25 & 94.75& & \\
\hline
\end{tabular}
\end{center}
\end{table}

\begin{table}[!ht]
\caption{Cross validation classification accuracy (\%) of middle phalange III Tanner-Whitehouse stage on the training set.}
\label{tab:TWS_MP3}
\begin{center}
\scriptsize
\begin{tabular}{| r | c c | c c | c c |}
\hline
           & \multicolumn{2}{c|}{Shape Features} &  \multicolumn{2}{c|}{Shapelet Transform} & \multicolumn{2}{c|}{Elastic Ensemble}

            \\ \hline
Classifier          & \textbf{Correct} & \textbf{Within One} & \textbf{Correct } & \textbf{Within One} & \textbf{Correct } & \textbf{Within One}\\
\hline
\hline
 \textbf{NN} & 75.25 & 98.75          & 55.39 & 86.68 & 64.66     & 87.47\\
 \textbf{Naive Bayes} & 77.25 & 99.25   & 64.66 & 89.20 & & \\
 \textbf{C4.5} & 69.50 & 98.00          & 54.89 & 79.15& & \\
 \textbf{SVML} & \textbf{77.75} & 98.50 & 59.18 & 86.43 & & \\
 \textbf{SVMQ} & 75.75 & \textbf{99.75} & 59.14 & 87.19& & \\
 \textbf{Rand. Forest} & 77.25 & 99.50 & 62.90 & 90.95& & \\
 \textbf{Rot. Forest} & 75.00 & 99.25  & 60.15 & 88.44 & & \\
\hline
\end{tabular}
\end{center}
\end{table}

\begin{table}[!ht]
\caption{Cross validation classification accuracy (\%) of proximal phalange III Tanner-Whitehouse stage on the training set.}
\label{tab:TWS_PP3}
\begin{center}
\scriptsize

\begin{tabular}{| r | c c | c c | c c |}
\hline
           & \multicolumn{2}{c|}{Shape Features} &  \multicolumn{2}{c|}{Shapelet Transform} & \multicolumn{2}{c|}{Elastic Ensemble}

            \\ \hline
Classifier          & \textbf{Correct} & \textbf{Within One} & \textbf{Correct } & \textbf{Within One} & \textbf{Correct } & \textbf{Within One}\\
\hline
\hline
 \textbf{NN} & 80.25 & 99.00          & 72.50 &   98.25  & 81.00 & 98.25\\
 \textbf{Naive Bayes} & 78.75 & 99.75   & 79.75 &   99.25 & &\\
 \textbf{C4.5} & 81.50 & 97.75          & 73.75 &   97.74 & &\\
 \textbf{SVML} & 82.50 & 98.75          & 80.00 &   99.75 & &\\
 \textbf{SVMQ} & \textbf{87.25} & \textbf{100.00} &  79.50   & 99.75 & &\\
 \textbf{Rand. Forest} & 85.75 & 98.75 &    81.25   &     99.50  & &\\
 \textbf{Rot. Forest} & 85.00 & 99.00 &     79.75   &   99.25  & &\\
\hline
\end{tabular}
\end{center}
\end{table}

Generally, all the classifiers are more accurate on the proximal phalange and less accurate on the middle phalange, but are nearly always within one class on all data. SVMQ is the most accurate classifier for the shape features and random forest for the shapelets and hence are the models we select from training. The elastic ensemble is designed to work with a nearest neighbour classifier only. The test results are shown in Table~\ref{tab:TWS_test}. The results are broadly consistent with the training results.

\begin{table}[!htb]
\caption{Testing Accuracy of SVMQ on Shape Features, Random Forest on Shapelets and the elastic ensemble}
\scriptsize
\label{tab:TWS_test}
\begin{center}
\begin{tabular}{| r | c c | c c | c c |}
\hline
           & \multicolumn{2}{c|}{Shape Features} &  \multicolumn{2}{c|}{Shapelet Transform} & \multicolumn{2}{c|}{Elastic Ensemble}

            \\ \hline
Classifier          & \textbf{Correct} & \textbf{Within One} & \textbf{Correct } & \textbf{Within One} & \textbf{Correct } & \textbf{Within One}\\
\hline
\hline
 \textbf{Distal} & 74.82 & 98.56    &  68.35 & 92.09    & 69.06 & 95.68 \\
 \textbf{Middle} & 75.97 & 100.00   & 55.19 & 87.66     & 51.80 & 81.17\\
 \textbf{Proximal} & 78.05 & 99.51  & 79.02 & 99.02     & 69.42 & 97.57\\
\hline
 \textbf{Overall} & 76.51 & 99.40 & 68.67 & 94.14 & 63.73 & 91.98 \\
\hline
\end{tabular}
\end{center}
\end{table}

The shape feature SVMQ is significantly better than both the shapelet random forest and the nearest neighbour elastic ensemble on middle phalanges and overall (using a McNemar's test).

Table~\ref{tab:testing_SVMQ_confusion} shows the confusion matrix for shape feature SVMQ on the test data. The main source of confusion is between stages F and G. Given the subjective nature of the labelling, some confusion is inevitable, and the decision between labelling F and G is perhaps the hardest.
\begin{table}[!htb]
\caption{Overall confusion matrix for SVMQ on unseen test data } \label{tab:testing_SVMQ_confusion}
\scriptsize
\begin{center}
\begin{tabular}{ c | c | c | c | c  | c | c } \hline & \textbf{D} & \textbf{E} & \textbf{F} & \textbf{G} & \textbf{H} & \textbf{I} \\ \hline \hline \textbf{D} &39 &16 & 0 & 0 & 0& 0 \\ \hline \textbf{E} & 6 & 98 & 15 & 0 & 0 & 0\\ \hline \textbf{F} & 0 & 16 & 75 & 18 & 3 & 0\\ \hline \textbf{G}  & 0 & 0 & 12 & 13 & 13 & 0\\ \hline \textbf{H} & 0 & 0 & 0 & 3 & 0 & 0\\ \hline \textbf{I} & 0 & 0 & 0 & 0 & 15 & 156\\ \hline
\end{tabular}
\end{center}
\end{table}

To put the performance into context, it is worth comparing these results to those of previously published TW classifiers. Thodberg {\em{et al.}}~\cite{thodberg2009bonexpert}  perform a cross validation on 84 images. Niemeijer {\em{et al.}}~\cite{niemeijer2003assessing} split their data into a training set of 119 images and a testing set of 71 images. Niemeijer {\em{et al.}} report an accuracy of 73.2\% correct and 97.2\% within one stage on the distal phalange. Our results and those of Thodberg {\em{et al.}} are presented in Table~\ref{tab:comparison}.

\begin{table}[!htb]
\caption{A comparison of results of the shape features SVMQ to those previously published, with all perecentages rounded down to the nearest whole number} \label{tab:comparison}
\scriptsize
\begin{center}
\begin{tabular}{c | c  | c | c} \hline
        & SVMQ Cross Validation & ASMA Test & Thodberg {\em{et al.}}~\cite{thodberg2009bonexpert} \\ \hline
 \textbf{Distal} & 84\%  (98\%)   & 74\% (98\%)      &   71\% (96\%)         \\
 \textbf{Middle} & 75\%  (99\%)    & 75 \% (100\%)    &   75\% (98\%)     \\
 \textbf{Proximal} & 87\%  (100\%) & 78 \% (99\%)     &   77\% (99\%)     \\
\hline
\end{tabular}
\end{center}
\end{table}

The data and experimental regime used to obtain these results are not the same, so we should be cautious in drawing any conclusions about relative performance. However, it seems that the three algorithms are broadly comparable, with approximately 75\%-80\% of cases correct and 95\%-100\% within one class. The results are also comparable to those of human raters as presented in~\cite{wenzel9skeletal}, where different observers rating the same radiograph, gave the same rating on 75-85\% instances and were within one stage on all instances.

Given the significantly higher accuracy of the shape feature classifier, we conclude this is the best of the approaches used for this problem.
Another benefit of using shape features is it allows an exploratory analysis of how the classifications are formed. By examining the information gain of each feature, we observe that the features measuring phalanx height (features 2 and 4) are important, but become less so when the epiphysis is present, when the width of the epiphysis (features 17 and 19) becomes more discriminatory.  Metaphysis measures (features 8 and 15) are more important for the distal phalange than the middle or proximal.  The reason for this is that  the middle and proximal phalanx are more tubular than the distal phalange and hence, there is less variation throughout development (feature 8)  as well as when in comparison to the width of the phalanx (feature 15), this is confirmed by the example images used for each TW stage~\cite{tanner-assessment}.

\section{Regressing onto Chronological Age}
\label{regression}
For clarity and simplicity, we restrict ourselves to the family of linear regression models for modelling age as a function of bone shape features. Linear regression has the advantage of producing models that are comprehensible and compact. However, care should be taken to ensure that the basic assumptions of normality, independence and constant variance hold. In Section~\ref{modelSelection} we describe the validation checks, feature selection and transformation stages we employ to ensure that the assumptions hold. In Section~\ref{prediction} we evaluate the predictive power of the regression models. This evaluation is performed through a leave one out cross validation (i.e. we do not assess predictive power with data used to construct the model). We compare predicted values to  chronological age and assess the model error in relation to the error for manual Greulich-Pyle bone age estimates made by two clinicians~\cite{gertych2007bone}. In Section~\ref{exploration} we employ the models to assess the relative importance of the features and compare models on different populations.

\subsection{Model Selection}
\label{modelSelection}

The core model selection technique we employ is stepwise forward selection regression, including all two variable interactions, using Akaike information criterion as the basis for the stopping condition.

We construct separate models for the instances where the epiphysis is detected and where it is not. We denote the models for the proximal phalange as $P_e,P_p$, where $P_e$ is the model constructed on data where the epiphysis is detected and $P_p$ is the model constructed on just the phalanx features. Similarly, the models built on just the distal phalange are denoted $D_e,D_p$ and the middle phalange are $M_e,M_p$.

In addition to examining regression models on single bones, we investigate ways of forming predictions from multiple bones. There are two obvious ways of doing this: we could either concatenate features and build the model on the expanded feature set, or we can produce a model for each bone and combine the predictions. We chose to combine estimates from individual bones. The main benefit of adopting this approach is that it is more flexible for cases when we cannot extract all the required bones. We denote models using the average of all the bones present as {\em DMP}. A further benefit of producing individual estimates for each bone is that it allows us to detect when the predictions are widely divergent and hence to detect outliers.

Linear regression is particularly susceptible to outliers since they can exert excessive leverage on the model. There are several unusual observations in our data. For example,  for the model $P_e$, there are three observations with a standardised residual greater than 2.5. An examination of the Cook's distance indicates that these three values are significantly influential to warrant investigation. The observed discrepancy between predicted and actual age may be caused by three factors. Firstly, the model may simply not capture the factors influencing age. Secondly, the child's development may be abnormal, meaning there is a genuine difference between bone age and chronological age. Thirdly, the checks we make to detect that the image processing has correctly captured the features may have failed. Examination of the images indicates that we have extracted the features correctly, and we have no way of knowing if the bone development of these children is normal or not.  Whilst the models would improve if we remove this data, it would also bias our evaluation of predictive power to do so. Instead, we mitigate against outliers for the combined model {\em DMP} by ignoring any prediction that is less than or great than 2 years of the other two predictions. This approach is unsupervised and hence will not bias our assessment of predictive power.

One of the core regression assumptions is that the variance of the errors is constant. To check this assumption, we measure the correlation between the absolute values of the standardised residuals and the response variable. If there is significant correlation, the assumption of constant variance is violated, and a transformation may be required.

\begin{figure}[!htb]
 		\begin{center}
		\subfigure[]{
		\includegraphics[height = 4cm]{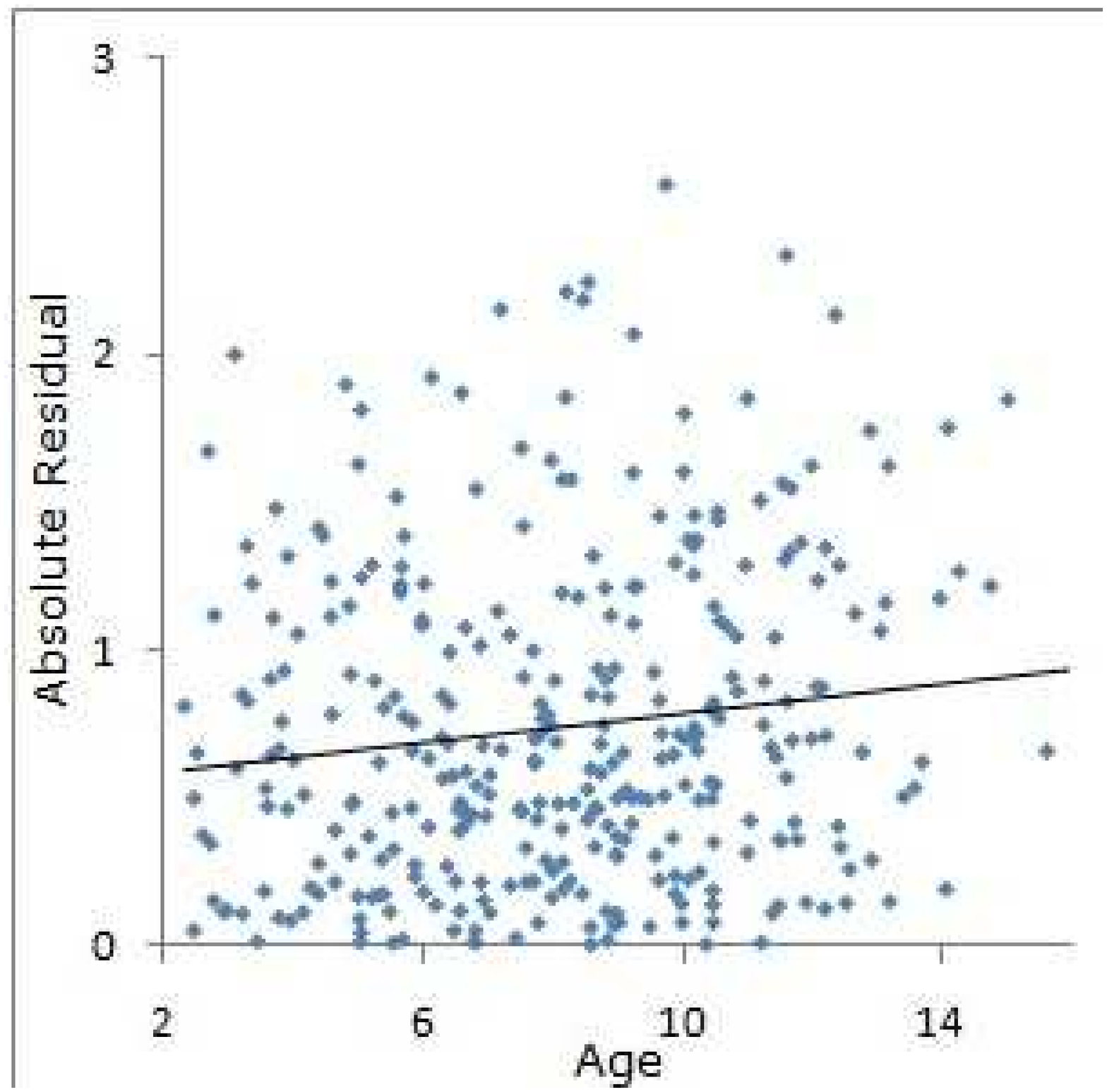}
		\label{img:res1}}
		\hspace{-0.1cm}
		\subfigure[]{
		\includegraphics[height = 4cm]{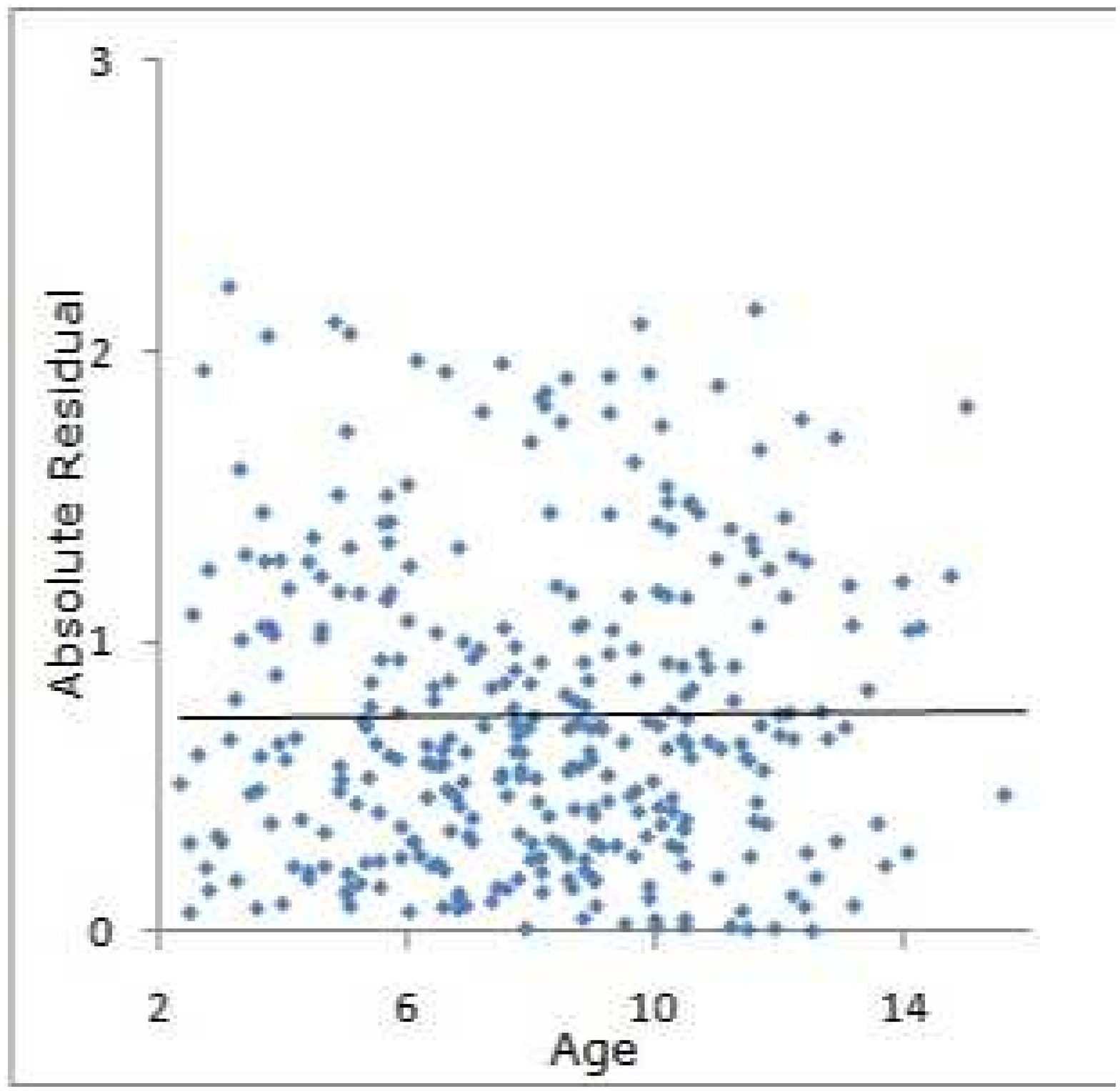}
		\label{img:res2}}
		\caption{Absolute standardised residuals against chronological age for the proximal phalanx with epiphysis~\subref{img:res1} prior to transform and~\subref{img:res2} after transform. The solid lines represent the regression of absolute residuals to age \label{absResPe}}	
		\end{center}

\end{figure}

Figure~\ref{absResPe}~\subref{img:res1} shows the plot of absolute standardised residuals against age for the $P_e$ model. The fitted linear regression line between the variables demonstrates the correlation. We find the same significant correlation on all three epiphysis models on all the cross validation subsets of images.

Changing variance, or heteroscedasity, is commonly dealt with through a power transform of the response. The Box-Cox procedure raises the response to the power $\lambda$, where $\lambda$ is selected to maximize likelihood. For the epiphysis models, choosing $\lambda$ in this way on the cross validation data sets yields a maximum likelihood transform value in the range 0.55-0.7. For simplicity, we will use a power transform value of 0.67 for all experiments (a value significant for all the data folds we experimented with). Figure~\ref{absResPe}~\subref{img:res2} shows how the transformation stabilises the residual variance.

For the non epiphysis models, we do not see the same pattern in the residuals. Instead, there is a consistent over estimating at the lower age range and underestimating for higher ages. This pattern of error is generally indicative of lack of predictive power. The Box-Cox transform (after subtracting 10 from the response to remove the scale effect) yields maximum likelihood $\lambda$ values are in the range 0.97 to 1.2, indicating that transforming the response will not help. Further experimentation with generalised linear modelling and with regressor transformations using the Box-Tidwell procedure did not improve the model. Hence, we do not transform the non epiphysis data. We discuss the usefulness of the non epiphysis models in Section~\ref{prediction}.

A further assumption of the linear model is normality. For the epiphysis models with the transformed response, we cannot reject the null hypothesis that the standardised residuals are normally distributed when performing the Shapiro-Wilk test of normality, D'Agostino's test of skewness and Jarque-Bera kurtosis test. We are confident that the linear regression assumptions hold for the epiphysis models with age transformed by raising it to the power 0.67.

However, the no epiphysis models $P_p, M_p$ and $D_p$ all fail the order statistic based Shapiro-Wilk test and the Jarque-Bera kurtosis test. The residual distribution is not skewed, but it does have higher kurtosis than it would if the normality assumption held. This is a result of the consistent under and over predicting caused by lack of predictive power.

\subsection{Predicting Age}
\label{prediction}

\begin{figure}[htb]
 		\begin{center}
		\includegraphics[height = 4cm]{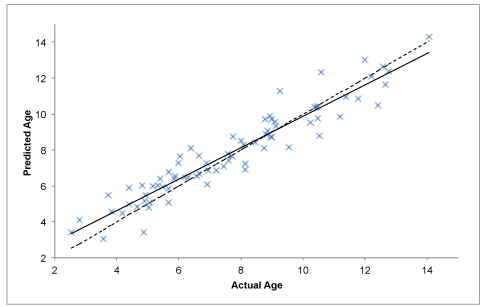}
		\caption{Predicted ages vs actual age for the model {\em DMP} with three bones present. The dotted line represents predicted equals actual. The solid line is the regression of predicted vs actual. \label{epiPred}}	
		\end{center}
\end{figure}

Table~\ref{tab:rmseEpi} shows the leave one out cross validation root mean square error (RMSE) of the linear models built on the individual bone features, and combinations of individual bone predictions, when the epiphysis is present. These results are presented in comparison to the RMSE of the scores given by clinicians using the GP system~\cite{gertych2007bone}. All models are constructed on age transformed by raising it to the power 0.67, but the RMSE scores are calculated by first transforming back to an age prediction. The results for the epiphysis bones are very encouraging. Increasing the number of bones in the model incrementally decreases the RMSE to the point where the three bone model is as accurate as expert human scorers. Figure~\ref{epiPred} plots the predicted age against the actual age for the {\em DMP} epiphysis model for cases when we have all three bones. There is a slight bias of under predicting young subjects and over predicting older subjects, but the {\em DMP} explains approximately 90\% of the variation in the response variable (based on the coefficient of determination, $R^2$).

\begin{table}[!htb]
\caption{RMSE Error for regression models where the epiphysis is detected. GP1 and GP2 are the RMSE for the two clinical estimates. \label{tab:rmseEpi}}
\scriptsize
 \begin{center}
\begin{tabular}{c | c | c | c | c} \hline
Model	          & Nos Cases &	Regression    & 	GP1      &	GP2 \\ \hline
\multicolumn{5}{c}{single bone models} \\ \hline
$D_e$	          &   275	  &  1.24	      &   0.89	     &  0.86 \\
$M_e$	          &   335     &	 1.27	      &   0.85	     &  0.92 \\
$P_e$	          &   334	  &  1.12	      &   0.87	     &  0.86 \\ \hline
\multicolumn{5}{c}{Multiple bone models {\em DMP}} \\ \hline
At least 1 bone   &   566	  & 1.19	      & 0.87	     & 0.89 \\
At least 2 bones  &	  294	  & 1.03	      & 0.86	     & 0.87 \\
3 bones	          &   76	  & 0.88	      & 0.89	     & 0.89 \\ \hline
\end{tabular}
\end{center}
\end{table}

\begin{figure}[htb]
 		\begin{center}
		\includegraphics[height = 4cm]{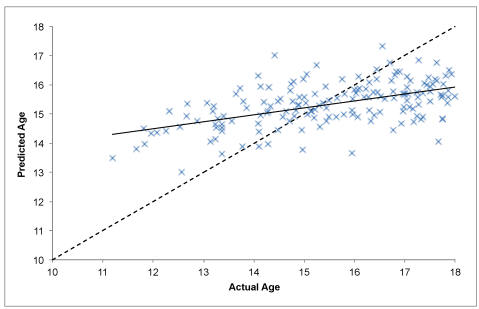}
		\caption{Predicted ages vs actual age for the no epiphysis model {\em DMP} with three bones present. The dotted line represents predicted equals actual. The solid line is the regression of predicted vs actual. \label{fig:noEpiPred}}	
		\end{center}
\end{figure}

Table~\ref{tab:rmseNoEpi} shows the results for bones when the epiphysis is not detected. Although the error decreases as bones are added to the model, the combined {\em DMP} model is still less accurate than the human scorers. {\em DMP} only explains approximately 28\% of the variation in age. Figure~\ref{fig:noEpiPred} plots the predicted age against the actual age for the {\em DMP} model for cases with no epiphysis where we have three bones. There is a consistent trend of overestimating the age of younger patients and underestimating old patients.

\begin{table}[!htb]
\caption{RMSE Error for regression models where the epiphysis is not detected. GP1 and GP2 are the RMSE for the two clinical estimates. \label{tab:rmseNoEpi} }
\scriptsize
\begin{center}
\begin{tabular}{c | c | c | c | c} \hline
Model	          & Nos Cases &	Regression    & 	GP1      &	GP2 \\ \hline
\multicolumn{5}{c}{single bone models} \\ \hline
$D_p$	          &  261	  &   1.56       &  1.24    &  1.29 \\
$M_p$	          &  217     &    1.48	     &  1.22	&  1.24 \\
$P_p$	          &  267	  &   1.6        &  1.19    &  1.24 \\ \hline
\multicolumn{5}{c}{Multiple bone models {\em DMP}} \\ \hline
At least 1 bone   &   320 	  & 1.53      & 1.22     & 1.26\\
At least 2 bones  &	  257 	  & 1.48      & 1.24     & 1.28\\
3 bones	          &   165     & 1.43	      & 1.17	     & 1.19 \\ \hline
\end{tabular}
\end{center}
\end{table}

There are several reasons why the no epiphysis models are worse than the epiphysis models. Firstly, predicting age for subjects approaching full maturity is generally much harder. After development stops bone features are no longer predictive of age, and the age at which development stops is highly variable. Another factor is that, based on the TW descriptors, intensity information is more important in distinguishing between almost fully mature bones. Including image intensity features may reduce the error. Finally, the assumptions of the linear model are clearly violated, hence an alternative modelling technique may reduce the error. We have experimented with ridge regression, response and regressor transforms and generalised linear modelling (with a variety of link functions), but have found no significant improvement.  Our approach in the future will be to combine the TW classification stage to filter out those cases that are fully matured. Our preliminary investigations indicate that this greatly improves the error for bones without an epiphysis. However, it is far more common to use bone ageing with subjects who are not fully mature, and our models indicate that the ASMA system, using just three bones, can predict as accurately as human experts.

The only published results we have been able to find that compare predicted age to actual age are in Adeshina {\em{et al.}}~\cite{adeshina-evaluating}, who report a Mean Absolute Error (MAE) for a Distal$+$Middle regression model of 1.26 for females and 1.28 for males. Table~\ref{tab:mae} presents the MAE for the comparable epiphysis models, non epiphysis models and the combined cases. These results demonstrate that even models built on a single bone without an epiphysis perform comparably to those reported in~\cite{adeshina-evaluating}, and when we have cases with and without the epiphysis, we perform much better.

\begin{table}[!htb]
\caption{Mean Absolute Error (MAE) for alternative bone combinations} \label{tab:mae}
\scriptsize
\begin{center}
\begin{tabular}{c | c | c | c }
 \hline
Model   &   Epiphysis	 &	            No-Epiphysis   &         Combined		\\
\hline
distal	& 1.01 	&   1.27  	&	1.14\\	
middle	& 0.98 	& 	1.23  	&  1.08	 \\
DM     & 0.91  &   1.16    &   1.05     \\
DMP	& 0.69  &  	1.2    &   1.04 \\
\hline
\end{tabular}
\end{center}
\end{table}

\vspace{-1cm}				
\subsection{Explanatory Analysis}
\label{exploration}

One of the benefits of adopting a linear regression model is the ease with which we can perform an exploratory analysis of the feature relevance. The epiphysis models include 15-17 terms, including a large number of interactions. The model for the proximal phalange is
\begin{eqnarray*}
age^{0.67}&=&-24.7+194 x_{19} + 0.2 x_{21}+19.4 x_{23}+12.5 x_{12}\\
& & +0.53 x_{22}-42.2  x_{15}-29.3 x_{20}-0.12 x_{16}\\
& &-0.12 x_{21}\cdot x_{15}-3.37  x_{23}\cdot x_{24} -9.4  x_{20} \cdot x_{24} \\
& &-0.13  x_{12} \cdot x_{22}+ 67x_{15}\cdot x_{20} -0.01   x_{24}\cdot x_{18}.
\end{eqnarray*}
The index of each variable corresponds to the feature number given in Table~\ref{tab:features}. Our first observation is that the model consists almost exclusively of epiphyseal features (the exception is $x_{15}$, metaphysis to width ratio). This is true for the other two bones also. This implies that future image processing efforts should focus more on accurately extracting and summarising the epiphysis.

Secondly, the features epiphysis width ($x_{19}$) and epiphysis distance to phalanx ($x_{21}$) are common to all models and are the first to enter the stepwise forward selection. Clearly they are the most important factors, and alone account for approximately 80\% of the variability in Age ($R^2$ on two variable linear model). The model constructed on just the proximal phalangeal features, epiphyseal width and epiphyseal distance to phalanx has a MAE of 0.98 and RMSE of 1.67. This implies that it would be very easy to construct a simple, practical model that would give a fairly accurate estimate based on two measurements that can be quickly performed by a non-specialist directly from the image. This offers the potential for screening for abnormality at very low cost.

A further benefit of the linear model is that, if the regression assumptions hold, we can construct confidence and prediction intervals for new data. This improves utility of the model in the decision making process because the decision of whether development is abnormal can be phrased as a hypothesis test where the null hypothesis is that the difference between bone age and chronological age is zero.

A linear model also offers a simple way of determining whether there are differences in age model between populations. We address the question of whether the models for male and female are significantly different by adding a factor to indicate sex. With regressors $x_{19}, x_{21}$ and $s$, where $s=0$ if the subject is male and $s=1$ if female, we find that we cannot reject the null hypothesis that the $s$ coefficient is zero and the resulting model is
\begin{eqnarray*}
age^{0.67}=0.246+0.02 x_{19}+ 0.02 x_{21} -0.11 s.
\end{eqnarray*}
If we fit a stepwise model, sex is the fourth variable to enter the model, and the interaction with $x_{19}$ is also significant. Clearly, sex is a predictive variable and future models should include the term.

The other demographic variable we have available is ethnicity. To test the significance of ethnicity, we include for factors to model whether the subject was Asian (a), African-American (f) or Hispanic (h). The only significant factor we find is whether subject is Asian. This is significant in the simplified model and in the stepwise model it is the second most important variable. Clearly there is a different development process at work for the patients with Asian ethnicity used in this study. If we include both sex and ethnicity, the {\em DMP} epiphysis model with three bones has a RMSE of 0.855 (compared to the human raters whose estimates had RMSE of 0.89).

\section{Conclusions and Future Work}
\label{sec:conclusions}

We describe three alternative approaches for using bone outlines to classify bone age stage and conclude that the shape features based on TW descriptors is the most appropriate. We then use these shape features to construct regression models of chronological age. The results for models predicting both Tanner-Whitehouse stage and chronological age are at least as good as those reported for other automated bone ageing systems. Furthermore, with just three bones, we produce age estimates that are as accurate as expert human assessors using the whole image. Data and code for each of these classification problems is available from~\cite{BoneWeb}.

In addition to the predictive accuracy, there are several benefits of using feature based regression models. Firstly, we can explain the importance of individual variables. Just two variables account for about 80\% of the variation in age, and this relationship offers the potential for fast screening for abnormalities. Secondly, when the regression assumptions are valid, we can construct confidence and prediction intervals. This can aid diagnosis and implies that increasing the training set size will incrementally improve the models (by reducing the variance). Finally, we can test the importance of alternative demographic variables and construct models tailored to specific populations. We demonstrate this  through an assessment of sex and ethnicity to the model. This offers several interesting possibilities: film free hospitals could enhance the quality of the general model through including their own data; geographic and demographic effects on bone development can be studied; historic data could be mined to quantify the effects of development drugs. There are many potential applications for an accurate age model constructed on a diverse and expanding database of images.

There are several obvious ways of improving our models. We shall include more bones and examine the effect of intensity features. We can investigate alternative segmentation and outline classification algorithms. We can attempt to screen for full maturity to improve the no epiphysis models. We can estimate age via the full TW methodology rather than directly. Our conclusion from the research we have conducted to date is that the feature based system of separating the image processing from the age modelling is the best approach. It offers flexibility, transparency and produces accurate estimates.

\end{document}